\documentclass{article} 
\usepackage{sure_preprint,times}

\usepackage{amsmath,amsfonts,bm}

\def\eqref#1{equation~\ref{#1}}

\def\1{\bm{1}}

\DeclareMathAlphabet{\mathsfit}{\encodingdefault}{\sfdefault}{m}{sl}
\SetMathAlphabet{\mathsfit}{bold}{\encodingdefault}{\sfdefault}{bx}{n}

\usepackage{xcolor}
\usepackage{graphicx}
\usepackage{float}
\usepackage{booktabs}
\usepackage{pifont}
\usepackage{xurl}
\usepackage{hyperref}
\usepackage{multirow}
\usepackage{tabularx}
\hypersetup{
  pdftitle={SURE: Spatiotemporal Uncertainty Representation for Efficient Vision-Language-Action Learning},
  pdfauthor={Jingye Zhang, Choo Sin Wai, Dake Zhong, Zhifang Liu, Yuhong Zhang, Siyao Xiao, Guorui Song, Guocun Wang, Dongbin Zhang, Peng Jiao, Haoqian Wang}}

\title{SURE: Spatiotemporal Uncertainty Representation for Efficient Vision-Language-Action Learning}

\author{%
\begin{minipage}{\textwidth}\centering\bfseries\small
Jingye Zhang$^{*}$ \enspace Choo Sin Wai$^{*}$ \enspace Dake Zhong$^{*}$ \enspace Zhifang Liu\\[0.3ex]
Yuhong Zhang \enspace Siyao Xiao \enspace Guorui Song \enspace Guocun Wang\\[0.3ex]
Dongbin Zhang \enspace Peng Jiao \enspace Haoqian Wang$^{\dagger}$\\[1ex]
\normalfont\normalsize Tsinghua University\\[0.5ex]
\footnotesize $^{*}$ Equal contribution. \quad $^{\dagger}$ Corresponding author.\\[0.2ex]
\texttt{jingye-z25@mails.tsinghua.edu.cn}
\end{minipage}%
}

\begin{document}

\maketitle

\begin{abstract}
   Vision-language-action (VLA) policies have become a promising paradigm for end-to-end robot control, but their effectiveness still depends heavily on large volumes of demonstration data. In manipulation tasks, this dependence is compounded by structured spatiotemporal uncertainty in the action space: multiple nearby actions or trajectories may successfully solve the same task, while semantically equivalent demonstrations often unfold at different speeds. Standard pointwise supervision treats each demonstrated action as an exact target, imposing an unnecessary learning burden under such uncertainty. We present SURE, a data-efficient VLA training framework that explicitly models both spatial and temporal uncertainty. SURE augments a VLA backbone with a lightweight transposed-convolutional probabilistic action decoder that captures spatial uncertainty through local correlations among neighboring action bins. It further introduces Soft-DTW alignment supervision to accommodate local timing variations between predicted and demonstrated action chunks. Together, these two components reduce sensitivity to demonstrator-specific poses and execution timing, enabling more data-efficient action learning. We evaluate SURE on LIBERO, RoboTwin 2.0, and real-world manipulation tasks. Even with only 10\% of the demonstration data or 20\% of the training steps, SURE achieves manipulation success rates that match or surpass baseline VLAs.
\end{abstract}

\begin{figure}[t]
    \centering
    \includegraphics[width=\textwidth]{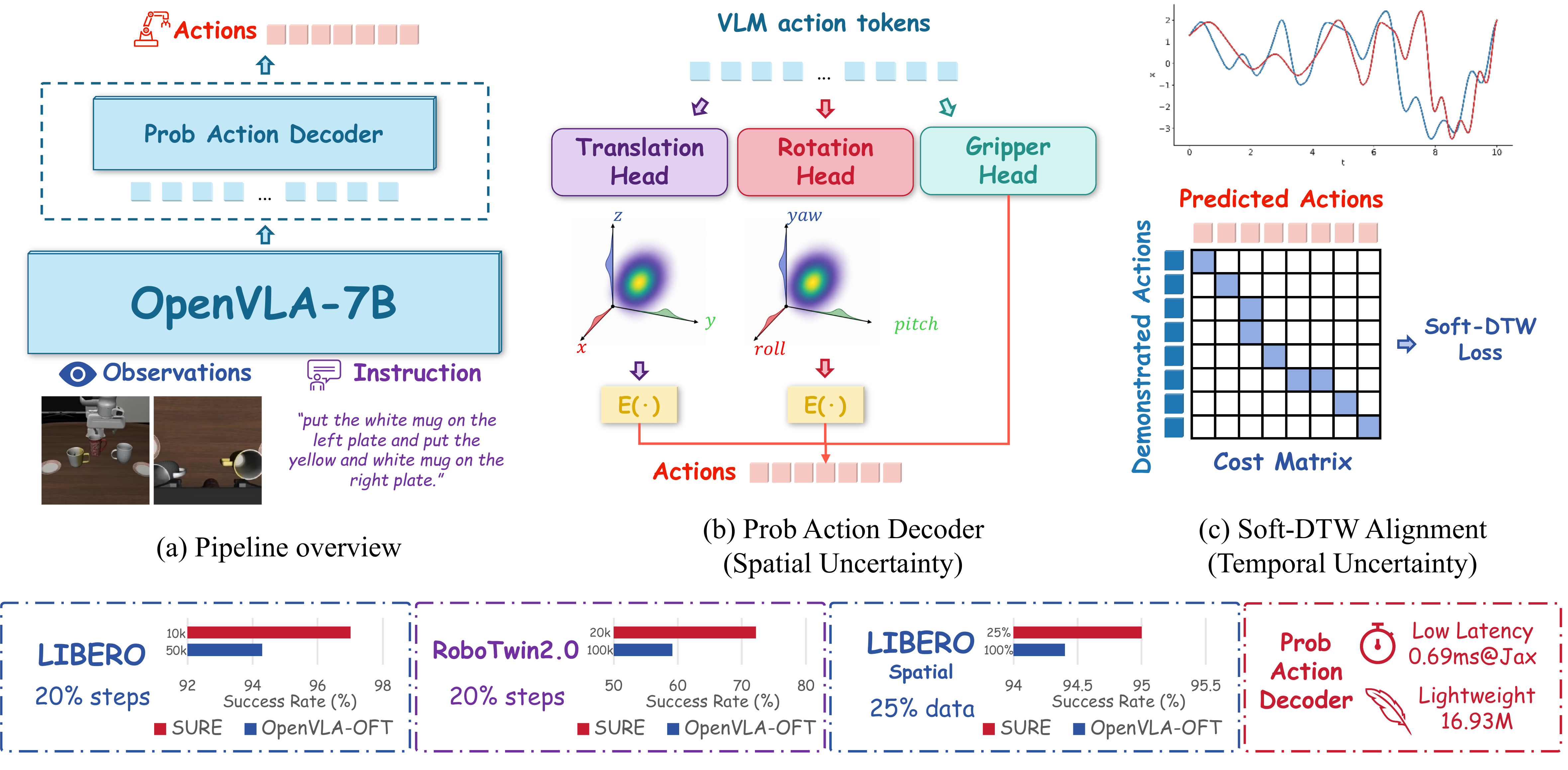}
    \caption{Overview of SURE. (a) Given observations and a language instruction, an OpenVLA-7B backbone produces action tokens that are decoded into action chunks. (b) Our probabilistic action decoder models translation and rotation as discretized distributions and recovers continuous controls with expectation-based decoding, capturing spatial uncertainty and multimodality. (c) During training, Soft-DTW alignment supervision aligns predicted and demonstrated action chunks that progress at different speeds, addressing temporal uncertainty in robot demonstrations.}
    \label{fig:intro_image}
\end{figure}

\section{Introduction}

Embodied AI and robot foundation models have advanced rapidly in recent years, driven by the emergence of large-scale vision-language-action (VLA) policies that directly map visual observations and language instructions to low-level control \citep{brohan2022rt1,ghosh2024octo,kim2024openvla,black2024pi0}. As a result, a central question in robot learning is no longer whe\-ther such end-to-end policies can work, but which action-learning paradigm can most efficiently turn expensive robot demonstrations into robust manipulation behaviors. This question is important in practical settings, where collecting real-world robot data remains slow, costly, and inherently noisy.

One source of difficulty is the inherent uncertainty in robot action data. Many manipulation tasks admit multiple valid solutions which the end effector can approach from slightly different poses, move at different speeds, or follow different but equally successful trajectories. Importantly, this uncertainty is structured rather than arbitrary. It typically exhibits strong local spatiotemporal correlations, actions that are close to a successful action in action space, or adjacent in time to a successful motion pattern, are often still correct. Treating supervision as a single exact target therefore introduces an unnecessary learning burden. A policy can be semantically correct yet still incur a large penalty simply because its prediction does not match the demonstrator's exact pose or timing.

Existing action representations only partially address this issue. Early and widely used VLA decoders often rely on tokenized or autoregressive action prediction \citep{brohan2022rt1,kim2024openvla}, while recent systems increasingly adopt diffusion- and flow-based action heads to better model multimodal futures \citep{chi2023diffusionpolicy,black2024pi0,bjorck2025gr00t}. These generative action heads are expressive and have become a popular choice for modern VLAs. However, their flexibility does not automatically yield data efficiency. When supervision still treats each action label as an isolated point, the model must discover from scratch that neighboring actions and slightly shifted trajectories should receive similar credit. In practice, this often means that the policy needs more data before it can internalize the local structure of valid robot behavior.

Recent work suggests that structured probabilistic action representations can be more sample-efficient and interpretable than unstructured point prediction. Instead of regressing a single action target, these methods model uncertainty with discretized action spaces, heatmap-like predictions, or voxel-space probability distributions, thereby injecting locality and multimodality into the output space \citep{james2021c2f,shridhar2022peract,goyal2023rvt,yang2026actionmap}. However, most of this literature focuses primarily on spatial uncertainty, while temporal uncertainty is still typically handled with exact per-step supervision.

Temporal uncertainty creates a second challenge. Even for the same task and intent, demonstrations can unfold at different speeds because of operator preferences, hesitation, embodiment dynamics, or minor contact events. Standard framewise regression losses such as L1 or L2 fail to account for this uncertainty and instead penalize timing differences directly. As a result, the model may overfit to demonstrator-specific execution speeds rather than the underlying behavior. This problem is closely related to the alignment problem long studied in speech recognition, where dynamic time warping (DTW) was introduced to compare signals that differ in local speaking rate \citep{sakoe1978dtw}. Its differentiable variant, Soft-DTW \citep{cuturi2017softdtw}, offers a natural way to bring this idea into gradient-based policy learning, yet it remains underexplored as a supervision mechanism for action prediction in VLA-style robot policies.

In this paper, we tackle both spatial and temporal uncertainty with a simple, data-efficient VLA design, as shown in Figure~\ref{fig:intro_image}. We introduce a lightweight probabilistic action decoder built with transposed convolutions that models spatial uncertainty through local correlations among neighboring action bins and can be supervised directly with standard pointwise losses. We further couple this head with Soft-DTW alignment supervision, which aligns action sequences executed at different speeds and accounts for temporal uncertainty across demonstrations. Together, these design choices encourage the policy to learn a distribution over valid behaviors rather than memorize a one-to-one mapping from observations to exact action labels. Empirically, SURE outperforms strong baselines with matched or fewer optimization steps. On LIBERO\citep{liu2023libero}, our method reaches 97.0\% average success after 10k steps, surpassing OpenVLA-OFT trained for 50k steps. With only 25\% of the LIBERO-Spatial data, our method also exceeds the full-data baseline. These gains extend to RoboTwin 2.0 \citep{chen2025robotwin2} and our real-world benchmark. 

Our contributions are threefold. First, we introduce a lightweight transposed-convolutional probabilistic action decoder that models spatial uncertainty with locally correlated action bins while enabling fast, parallel expectation-based decoding without iterative denoising. Second, we incorporate Soft-DTW alignment supervision to improve robustness to demonstrations with similar task intent but different execution speeds. Third, extensive experiments on LIBERO, RoboTwin 2.0, and our real-world benchmark show that our method achieves stronger training and data efficiency than competitive baselines, requiring fewer optimization steps or less training data.


\section{Related Work}

\noindent\textbf{Embodied AI}
Large-scale embodied AI is currently pursued through two complementary paradigms. Vision-language-action (VLA) policies directly map observations and language instructions to executable actions, typically benefiting from large, heterogeneous robot datasets and pretrained multimodal backbones \citep{brohan2022rt1,oxe2023rtx,ghosh2024octo,kim2024openvla,black2024pi0,bjorck2025gr00t}. In parallel, world-model and world-action-model approaches augment policy learning with future prediction, imagination, or simulation, enabling planning, policy evaluation, or synthetic data generation in learned latent or pixel spaces \citep{hafner2019dreamer,liao2025genie,guo2025ctrlworld,zhou2026tau0wm}.

\noindent\textbf{Action Representation}
Action representation is a central design choice in robot policy learning. Early VLA and imitation-learning systems commonly relied on discretized or autoregressive action prediction, where continuous control is converted into token sequences or predicted one step at a time \citep{brohan2022rt1,kim2024openvla,pertsch2025fast}. More recent work has explored continuous parallel decoding and improved fine-tuning recipes to reduce inference cost and increase adaptation performance \citep{kim2025oft,wen2024tinyvla}. Another major direction models actions as distributions rather than single point estimates. This includes diffusion- and flow-based action heads for multimodal action generation \citep{chi2023diffusionpolicy,black2024pi0,bjorck2025gr00t}, as well as structured probabilistic outputs such as discretized spatial bins, heatmaps, and voxel-space distributions \citep{james2021c2f,shridhar2022peract,goyal2023rvt,yang2026actionmap}. Our method is most closely related to the latter line of work. Specifically, we use a transposed-convolutional head to explicitly model the action probability distribution and its decoding process. We optimize the decoder primarily with L1 action supervision and use only a lightly weighted auxiliary distribution loss, rather than relying on dense distributional supervision as the primary learning signal.

\noindent\textbf{DTW-Based Temporal Alignment}
Dynamic time warping (DTW) has long been a standard tool for aligning sequences with non-uniform temporal progress \citep{sakoe1978dtw}. Soft-DTW extends this idea to a differentiable objective, making it suitable for gradient-based learning \citep{cuturi2017softdtw}. In robotics, DTW-style alignment has been used for demonstration comparison, weak supervision, and imitation under temporal or action mismatches \citep{yeh2025acil}. These works highlight the practical importance of alignment under variable execution speeds when different trajectories express the same underlying intent. Rather than using DTW only for retrieval, comparison, or offline alignment, we use Soft-DTW alignment supervision for VLA action learning, allowing the policy to learn end-to-end from temporally misaligned but semantically consistent demonstrations.

\noindent\textbf{Data-Efficient Robot Learning.}
Improving data efficiency remains a core challenge in robot learning because large-scale demonstrations are expensive to collect, especially on real hardware. Existing efforts improve efficiency from several angles: scaling diverse cross-embodiment datasets for transfer \citep{oxe2023rtx,ghosh2024octo,kim2024openvla}, designing better adaptation and fine-tuning recipes \citep{kim2025oft}, building smaller or faster VLA architectures \citep{wen2024tinyvla}, and injecting stronger geometric structure into the action output space \citep{james2021c2f,shridhar2022peract,goyal2023rvt,yang2026actionmap}. Our approach is complementary to these directions. Rather than relying on more pretraining data, larger-scale models, or strongly weighted target-distribution supervision, we improve data efficiency through a joint design of structured action representation and temporally tolerant supervision.

\begin{figure}[t!]
    \centering
    \includegraphics[width=0.6\linewidth]{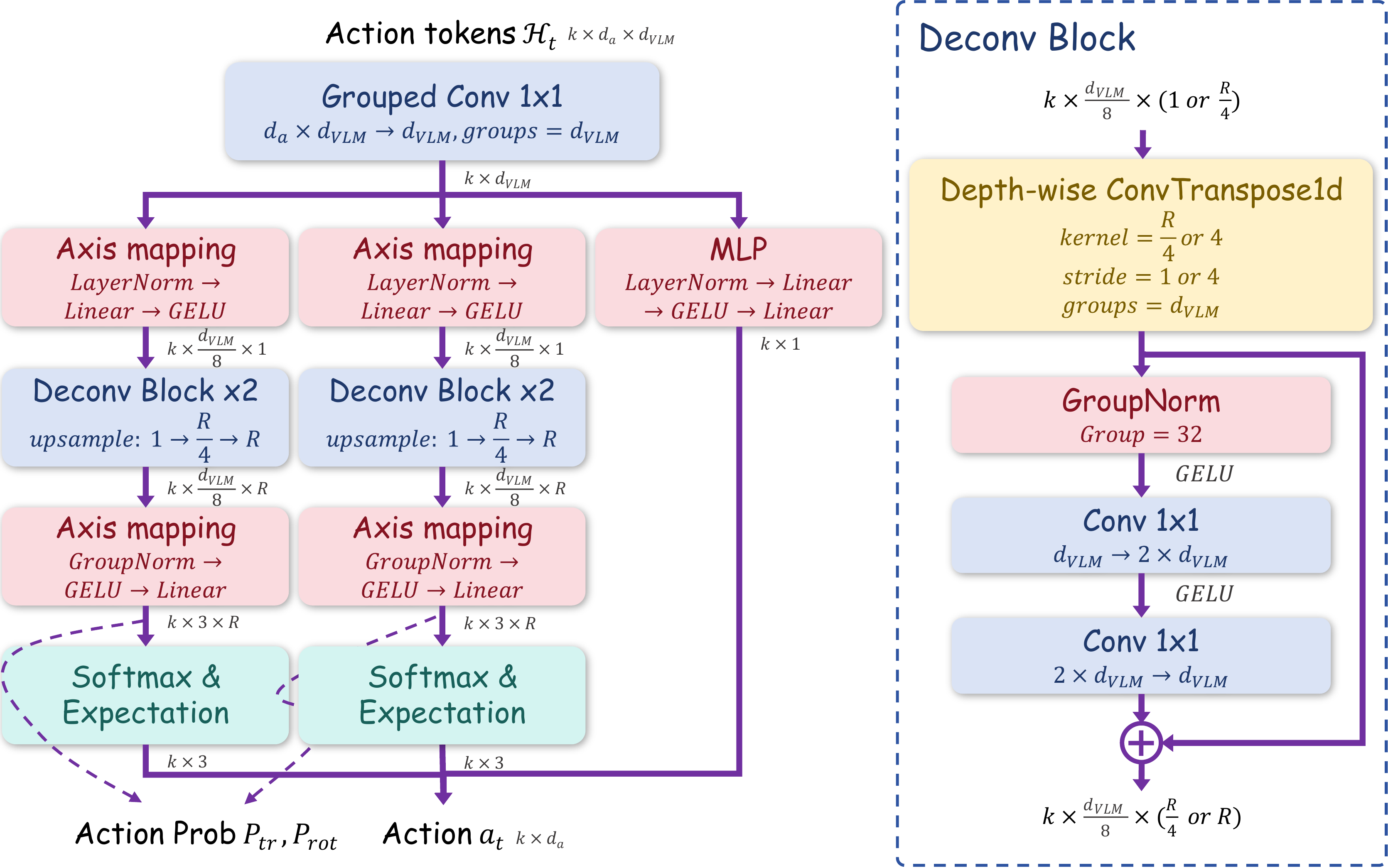}
    \caption{Architecture of the probabilistic action decoder. Action tokens $\mathcal{H}_t \in \mathbb{R}^{k \times d_a \times d_{VLM}}$ are first fused by a grouped $1 \times 1$ convolution and then split into translation, rotation, and gripper branches. Translation and rotation use two Deconv Blocks to upsample an ordered bin axis from $1 \rightarrow R/4 \rightarrow R$. The resulting per-axis logits over $R$ bins are normalized by softmax and decoded into continuous actions by expectation. Each Deconv Block combines depthwise transposed convolution with pointwise residual layers to encode local correlations among neighboring bins. The gripper action is decoded by a lightweight MLP.}
    \label{fig:ProbActionDecoder}
\end{figure}

\section{Methods}

Our method builds on an OpenVLA-7B backbone and introduces two uncertainty-aware components for data-efficient action learning. The probabilistic action decoder represents spatial uncertainty in translation and rotation, while Soft-DTW alignment supervision accounts for temporal uncertainty among semantically consistent action sequences. At time step $t$, the policy takes a visual observation $\mathcal{I}_t$, proprioceptive state $p_t$, and language instruction $\mathcal{T}$ as input. The VLA backbone produces a chunk of action tokens
\begin{equation}
    \mathcal{H}_t = \mathrm{VLM}(\mathcal{I}_t, p_t, \mathcal{T})
    \in \mathbb{R}^{k \times d_a \times d_{VLM}},
\end{equation}
where $k$ is the action chunk length, $d_a$ is the action dimension, and $d_{VLM}$ is the hidden size of the backbone. The probabilistic action decoder maps these tokens to an action chunk $\mathcal{A}_t \in \mathbb{R}^{k \times d_a}$. Soft-DTW alignment supervision then aligns predicted and demonstrated action chunks whose trajectories are semantically consistent but differ in local execution rate.

\subsection{Probabilistic Action Decoder}
Robot manipulation rarely admits a unique action target. Even under the same observation and instruction, multiple nearby end-effector poses, orientations, or contact timings can all lead to success. This makes the action space both locally correlated and potentially multimodal. Direct scalar regression ignores this structure and forces the model to match one exact demonstrator trajectory.

For each continuous translation and rotation dimension, we discretize the normalized action range $[-1,1]$ into $R=256$ ordered bins with centers $\{c_r\}_{r=1}^{R}$. As illustrated in Figure~\ref{fig:ProbActionDecoder}, the decoder first compresses $\mathcal{H}_t$ with a grouped convolution and then sends the resulting features to translation, rotation, and gripper branches. The translation and rotation branches use lightweight transposed-convolutional upsampling to produce logits
\begin{equation}
    \mathbf{Z}_{\mathrm{tr}}, \mathbf{Z}_{\mathrm{rot}} \in \mathbb{R}^{k \times 3 \times R}.
\end{equation}
The transposed-convolutional structure provides the key inductive bias. Because adjacent bins correspond to nearby continuous actions, convolutional weight sharing encourages neighboring bins to receive correlated evidence rather than be treated as unrelated classes. This gives the decoder a native mechanism for modeling spatial uncertainty characterized by local correlations among neighboring action bins.

The logits are normalized into categorical action probabilities
\begin{equation}
    \mathbf{P}_{\mathrm{tr}} = \mathrm{softmax}(\mathbf{Z}_{\mathrm{tr}}), \quad
    \mathbf{P}_{\mathrm{rot}} = \mathrm{softmax}(\mathbf{Z}_{\mathrm{rot}}),
\end{equation}
where the softmax is applied along the bin dimension. A continuous action value is obtained as the expectation over bin centers,
\begin{equation}
    \label{eq:action_expectation}
    a_{n,j} = \sum_{r=1}^{R} \mathbf{P}_{n,j,r} c_r,
\end{equation}
for chunk index $n$ and action axis $j$, where $\mathbf{P}$ denotes the corresponding translation or rotation probability tensor. Because this expectation is differentiable, L1 action supervision can optimize the decoder end to end. A lightly weighted auxiliary distribution loss further regularizes the predicted probabilities. For the gripper dimension, we use a small MLP rather than a transposed-convolutional distribution head, since binary or near-binary gripper commands have much weaker spatial uncertainty than translation and rotation.

\subsection{Soft-DTW Alignment}

Robot demonstrations can be semantically consistent while differing in local execution speed. For example, one demonstration may approach an object slowly and then close the gripper quickly, while another may execute the same behavior at a more uniform speed. Framewise L1 supervision penalizes such local timing shifts even when the action chunks describe the same underlying behavior. Inspired by dynamic time warping in speech recognition \citep{sakoe1978dtw}, we use Soft-DTW \citep{cuturi2017softdtw} as a differentiable alignment loss between predicted and demonstrated action chunks.

Let $\mathcal{A}_t=\{a_i\}_{i=1}^{k}$ and $\mathcal{A}^{\ast}_t=\{a^{\ast}_j\}_{j=1}^{k}$ denote the predicted and demonstrated action chunks. Excluding the final gripper dimension, we form cumulative trajectories $\mathcal{X}_t=\{x_i\}_{i=1}^{k}$ and $\mathcal{X}^{\ast}_t=\{x^{\ast}_j\}_{j=1}^{k}$, where $x_i=\sum_{\ell=1}^{i}a_{\ell,1:d_a-1}$ and $x^{\ast}_j=\sum_{\ell=1}^{j}a^{\ast}_{\ell,1:d_a-1}$. We define the pairwise trajectory cost as
\begin{equation}
    C_{ij}=\frac{1}{d_a-1}\lVert x_i-x^{\ast}_j\rVert_2^2,
\end{equation}
which measures the mean squared discrepancy between two candidate aligned steps. Soft-DTW replaces the hard minimum in DTW with a differentiable soft minimum. With $D_{0,0}=0$ and $D_{i,0}=D_{0,j}=+\infty$, we compute
\begin{equation}
    \label{eq:softdtw}
    \begin{aligned}
        D_{i,j} = C_{ij} + \operatorname{softmin}_{\gamma}(D_{i-1,j-1}, D_{i-1,j}, D_{i,j-1}), \\
        \operatorname{softmin}_{\gamma}(z_1,z_2,z_3) = -\gamma \log \sum_{m=1}^{3}\exp(-z_m/\gamma).
    \end{aligned}
\end{equation}
The temperature $\gamma$ controls the alignment smoothness, with smaller values approaching standard DTW. We denote the resulting cost by $D_{\gamma}(\mathcal{X}_t,\mathcal{X}^{\ast}_t)=D_{k,k}$.

To remove the self-similarity bias of the raw alignment score, we use the Soft-DTW divergence
\begin{equation}
        \mathcal{L}_{\mathrm{div}}=
        D_{\gamma}(\mathcal{X}_t,\mathcal{X}^{\ast}_t)
        -\frac{1}{2}D_{\gamma}(\mathcal{X}_t,\mathcal{X}_t) -\frac{1}{2}D_{\gamma}(\mathcal{X}^{\ast}_t,\mathcal{X}^{\ast}_t).
\end{equation}
This objective encourages the predicted cumulative trajectory to follow a temporally flexible monotonic alignment path to the demonstration. As a result, the policy can learn from semantically consistent action chunks even when their local timing differs.

\subsection{Supervision}

For spatial uncertainty, we primarily supervise the continuously decoded actions with
\begin{equation}
    \mathcal{L}_{\mathrm{L1}}
    = \frac{1}{k d_a}
    \sum_{n=1}^{k}\sum_{j=1}^{d_a}
    \left|a_{n,j}-a^{\ast}_{n,j}\right|,
\end{equation}
where $a$ and $a^{\ast}$ are the predicted and demonstrated actions. Differentiable expectation-based decoding allows this loss to optimize the probabilistic decoder end to end. Assuming the gripper is the last action dimension, we further impose a weak KL constraint on the remaining $d_a-1$ translation and rotation dimensions:

\begin{equation}
    \mathcal{L}_{\mathrm{KL}}
    = \frac{1}{k(d_a-1)}
    \sum_{n=1}^{k}\sum_{j=1}^{d_a-1}
    D_{\mathrm{KL}}\!\left(
    \mathbf{q}_{n,j}\,\|\,\mathbf{P}_{n,j}
    \right).
\end{equation}
Here, $\mathbf{P}$ is the predicted distribution and $\mathbf{q}$ is a discretized Gaussian centered at the demonstrated action, with standard deviation $\sigma$ as a bandwidth hyperparameter. The divergence is
\begin{equation}
    D_{\mathrm{KL}}(\mathbf{q}\,\|\,\mathbf{P})
    = \sum_{r=1}^{R} q_r \log \frac{q_r}{P_r}.
\end{equation}
We set $\lambda_{\mathrm{KL}}=0.05$ empirically, so KL serves only as a stabilizing regularizer while L1 remains the primary spatial supervision.

For temporal uncertainty, we combine the Soft-DTW divergence $\mathcal{L}_{\mathrm{div}}$ with an endpoint loss
\begin{equation}
    \mathcal{L}_{\mathrm{end}}
    = \left\|x_k-x^{\ast}_k\right\|_1,
\end{equation}
where $x_k$ and $x^{\ast}_k$ are the final cumulative trajectory points of the predicted and demonstrated chunks. Soft-DTW tolerates local timing uncertainty, while the endpoint term preserves final task progress.

The complete objective is
\begin{equation}
    \mathcal{L} = \mathcal{L}_{\mathrm{L1}} + \lambda_{\mathrm{KL}}\mathcal{L}_{\mathrm{KL}} + \lambda_{\mathrm{DTW}} \left( \mathcal{L}_{\mathrm{div}} + \mathcal{L}_{\mathrm{end}} \right).
\end{equation}
We set $\lambda_{\mathrm{DTW}}=0.5$ empirically. We optimize all loss terms jointly throughout training and anneal the Soft-DTW smoothing coefficient $\gamma$ in Eq.~\ref{eq:softdtw} from a larger initial value to its target value, transitioning from smooth alignment to more precise temporal matching.

\section{Experiments}\label{sec:experiments}
We evaluate SURE on the LIBERO and RoboTwin 2.0 simulation benchmarks and on six real-world manipulation tasks using demonstrations collected for this study. Implementation details for these evaluations are provided in Appendix~\ref{app:training-details}.

\subsection{Simulation Evaluation}

\subsubsection{LIBERO}

\noindent\textbf{Evaluation protocol.}
We evaluate SURE on the four standard LIBERO task suites, namely LIBERO-Spatial, LIBERO-Object, LIBERO-Goal, and LIBERO-Long. These suites collectively assess spatial reasoning, object-centric generalization, goal understanding, and long-horizon manipulation \citep{liu2023libero}. The baseline results are collected from sources that use different statistical protocols. All results marked with $\dagger$ in Table~\ref{tab:libero_result} are averaged over three random seeds, with 500 evaluation trials per task suite (50 trials for each of the 10 tasks). Specifically, the results for Diffusion Policy, Octo, and OpenVLA are taken from the OpenVLA evaluation \citep{kim2024openvla}. The GR00T-N1 and $\pi_0$ baselines are trained and reported by the authors of Hume \citep{song2026hume}. OpenVLA-OFT and SURE are evaluated using a single seed.

\noindent\textbf{Baseline methods.}
We compare SURE with a diverse set of representative policies spanning diffusion-based control, generalist robot policies, and VLA fine-tuning methods. For OpenVLA-OFT, we fine-tune the model using its publicly released implementation \citep{kim2025oft} and report results after both 10k and 50k optimization steps. SURE uses the same OpenVLA-7B backbone and is trained for only 10k optimization steps.

\noindent\textbf{Results.}
As shown in Table~\ref{tab:libero_result}, SURE achieves the highest average success rate of 97.0\% after only 10k optimization steps, surpassing OpenVLA-OFT by 9.9\% under the same training budget and achieving gains across all four task suites. SURE also surpasses the 50k-step OpenVLA-OFT model by 2.7\% in average success rate despite using only 20\% optimization steps. Across individual suites, SURE obtains the best performance on all four task suites, surpassing the baselines on LIBERO-Spatial, LIBERO-Object, LIBERO-Goal, and LIBERO-Long. These results indicate that explicitly modeling spatiotemporal uncertainty improves both final manipulation performance and training efficiency.

\begin{table}[htb]
    \centering
    \caption{Success rates (\%) on the four LIBERO task suites. Training steps and batch sizes (BS) are reported when available. SURE is trained for 10k optimization steps and achieves the highest average success rate. Bold values indicate the best result in each column. $\dagger$ denotes results averaged over three random seeds, all other results are obtained using a single evaluation seed.}
    \label{tab:libero_result}
    {
        \small
        \begin{tabular}{
            >{\arraybackslash}m{4.5cm}
            >{\centering\arraybackslash}m{1.9cm}
            *{5}{>{\centering\arraybackslash}m{0.9cm}}
            }
            \toprule
            Methods                    & Steps(BS) & Spatial       & Object        & Goal          & Long          & Avg.       \\
            \midrule
            Diffusion Policy$^\dagger$~\citep{chi2023diffusionpolicy} & 100-250k(64)       & 78.3          & 92.5          & 68.3          & 50.5          & 72.4          \\
            Octo$^\dagger$~\citep{ghosh2024octo}             & 100-250k(64)       & 78.9          & 85.7          & 84.6          & 51.1          & 75.1          \\
            OpenVLA$^\dagger$~\citep{kim2024openvla}          & 50-150k(64)        & 84.7          & 88.4          & 79.2          & 53.7          & 76.5          \\
            GR00T-N1$^\dagger$~\citep{bjorck2025gr00t}                   & /~(128)            & 94.4          & 97.6          & 93.0          & 90.6          & 93.9          \\
            $\pi_0$$^\dagger$~\citep{black2024pi0}                    & /~(128)            & 96.8          & 98.8          & 95.8          & 85.2          & 94.2          \\
            OpenVLA-OFT~\citep{kim2025oft}                & 10k(48)            & 94.4          & 92.6          & 92.6          & 68.8          & 87.1          \\
            OpenVLA-OFT                & 50k(48)            & 96.8          & 97.4          & 98.0          & 85.0          & 94.3          \\
            \midrule
            \textbf{SURE}               & 10k(48)            & \textbf{99.0} & \textbf{99.2} &  \textbf{98.8}          & \textbf{91.0}          & \textbf{97.0} \\
            \bottomrule
        \end{tabular}
    }
\end{table}

\begin{table}[htb]
    \centering
    \caption{Success rates (\%) on five RoboTwin 2.0 tasks under the Easy evaluation setting. All methods are trained on the clean subset with a separate checkpoint for each task. Training steps and batch sizes (BS) are reported for each method. Bold values indicate the best result in each column.}
    \label{tab:robotwin}
    {
        \small
        \setlength{\tabcolsep}{2pt}
        \begin{tabular}{
            >{\arraybackslash}m{2.8cm}
            >{\centering\arraybackslash}m{2.0cm}
            >{\centering\arraybackslash}m{1.45cm}
            >{\centering\arraybackslash}m{1.6cm}
            >{\centering\arraybackslash}m{1.05cm}
            >{\centering\arraybackslash}m{1.05cm}
            >{\centering\arraybackslash}m{1.6cm}
            >{\centering\arraybackslash}m{0.8cm}
            }
            \toprule
            Methods          & Training steps(BS) & Beat Block Hammer & Click Alarm Clock & Lift Pot & Place Fan & Stack Bowls Two & Avg. \\
            \midrule
            Diffusion Policy & 20-75k(128)          & 42.0          & 61.0           & 39.0          & 3.0           & 61.0          & 41.2          \\
            OpenVLA-OFT      & 100k(8)              & 18.0          & 85.0           & 93.0          & 25.0          & 75.0          & 59.2          \\
            RDT~\citep{liu2025rdt}              & 10k(64)              & \textbf{77.0} & 61.0           & 72.0          & 12.0          & 76.0          & 59.6          \\
            $\pi_0$          & 30k(32)              & 43.0          & 63.0           & 84.0          & 20.0          & \textbf{91.0} & 60.2          \\
            \midrule
            \textbf{SURE}     & 20k(16)              & 55.0          & \textbf{100.0} & \textbf{98.0} & \textbf{26.0} & 82.0          & \textbf{72.2} \\
            \bottomrule
        \end{tabular}
    }
\end{table}

\subsubsection{RoboTwin 2.0}

\noindent\textbf{Evaluation protocol.}
RoboTwin 2.0 is a large-scale bimanual manipulation benchmark containing 50 tasks and supporting multiple dual-arm embodiments \citep{chen2025robotwin2}. We evaluate on the five tasks listed in Table~\ref{tab:robotwin}: Beat Block Hammer, Click Alarm Clock, Lift Pot, Place Fan, and Stack Bowls Two. All methods are trained on the clean subset, which contains 50 clean expert demonstrations for each task. Following the single-task setting, we fine-tune or train a separate checkpoint for every task. We report the success rate over 100 evaluation rollouts per task under the Easy setting, which uses clean, non-randomized environments.

\noindent\textbf{Baseline methods.}
We compare SURE with Diffusion Policy, RDT\citep{liu2025rdt}, $\pi_0$, and OpenVLA-OFT. The results for Diffusion Policy, RDT, and $\pi_0$ are taken from the RoboTwin 2.0 paper \citep{chen2025robotwin2}. The OpenVLA-OFT results are taken from JEPA-VLA \citep{miao2026jepavla}. These baselines cover diffusion-based visuomotor control, pretrained vision-language-action policies, and efficient fine-tuning of an OpenVLA backbone.

\noindent\textbf{Results.}
As shown in Table~\ref{tab:robotwin}, SURE achieves the highest average success rate of 72.2\%, outperforming the strongest baseline, $\pi_0$, by 12.0\%. SURE obtains the best performance on three of the five tasks, reaching 100.0\% on Click Alarm Clock, 98.0\% on Lift Pot, and 26.0\% on Place Fan. It also achieves the second-best result on Beat Block Hammer and Stack Bowls Two. Compared with OpenVLA-OFT, SURE improves the average success rate by 13.0\% while using only 20k rather than 100k optimization steps. These results demonstrate that SURE improves both manipulation performance and training efficiency on bimanual tasks.

\subsection{Real-World Evaluation}

\noindent\textbf{Evaluation protocol.}
We evaluate SURE on six real-world manipulation tasks: Water Bottle, AAA Battery, Screwdriver, Toy Duck, Stack Bowls Two, and Pour Water. For each task, we collect 50 demonstration episodes through multiple authors acting as operators. Differences in operator motion and execution speed introduce natural spatial and temporal uncertainty into the demonstrations. Each policy is evaluated over 20 trials per task, and we report the corresponding success rate. Detailed descriptions of the task setups and execution procedures are provided in the Appendix\ref{app:real-world}.

\noindent\textbf{Baseline method.}
We compare SURE with OpenVLA-OFT using the same mixed-task real-world demonstrations. Both methods are jointly trained on demonstrations from all six tasks, and a single checkpoint is evaluated across all tasks. OpenVLA-OFT is trained for 50k optimization steps, whereas SURE is trained for only 20k steps.

\noindent\textbf{Results.}
As shown in Table~\ref{tab:real_world}, SURE achieves an average success rate of 57.5\%, outperforming OpenVLA-OFT by 15.8\% while using 40\% optimization steps. SURE performs better on five of the six tasks, with the largest improvement on Toy Duck, where the success rate increases from 45\% to 95\%. These results demonstrate that SURE learns effectively from real-world demonstrations with substantial spatial and temporal uncertainty.

\begin{table}[t!]
    \centering
    \caption{Success rates (\%) on six real-world manipulation tasks. Each method is evaluated over 20 trials per task. OpenVLA-OFT and SURE are trained for 50k and 20k optimization steps, respectively. Bold values indicate the best result in each column.}
    \label{tab:real_world}
    {
        \small
        \begin{tabular}{
            >{\arraybackslash}m{2.4cm}
            *{4}{>{\centering\arraybackslash}m{1.1cm}}
            >{\centering\arraybackslash}m{1.5cm}
            >{\centering\arraybackslash}m{1.1cm}
            >{\centering\arraybackslash}m{1.0cm}
            }
            \toprule
            Methods      & Water Bottle & AAA Battery & Screw\-driver & Toy Duck & Stack Bowls Two & Pour Water & Avg. \\
            \midrule
            OpenVLA-OFT  & 40           & 35              & 55          & 45          & 20          & \textbf{55} & 41.7          \\
            \textbf{SURE} & \textbf{55}  & \textbf{60}     & \textbf{70} & \textbf{95} & \textbf{25} & 40          & \textbf{57.5} \\
            \bottomrule
        \end{tabular}
    }
\end{table}

\subsection{Inference Efficiency Evaluation}

\noindent\textbf{Evaluation protocol.}
We evaluate the inference efficiency of the action head using models trained on LIBERO-Spatial, with the action chunk length fixed to $k=8$. To isolate the computational cost of action decoding, we measure only the action head or decoder and exclude the VLA backbone. Because $\pi_{0.5}$ \citep{intelligence2025pi05} is implemented in JAX, we additionally implement our action head in JAX to provide a framework-matched latency comparison. For each method, we report the number of action-head parameters, FLOPs, and mean chunk latency. Latency is averaged over 100 action-decoding runs following 20 warm-up runs.

\noindent\textbf{Results.}
As shown in Table~\ref{tab:inference_efficiency}, SURE contains only 16.93M action-head parameters and requires 10.96G FLOPs, corresponding to 3.9\% of the parameters and 13.8\% of the FLOPs of $\pi_{0.5}$. The standard SURE implementation generates an action chunk in 1.57\,ms, providing a $12.7\times$ speedup over $\pi_{0.5}$. Under the framework-matched JAX comparison, SURE reduces the latency to 0.69\,ms, corresponding to a $28.8\times$ speedup over $\pi_{0.5}$. Although the deterministic OFT head remains faster and requires fewer FLOPs, SURE uses only 11.2\% as many parameters while retaining an uncertainty-aware action representation. These results show that SURE provides efficient uncertainty-aware action decoding without the high inference cost of iterative generative action heads.

\begin{table}[htb]
    \centering
    \caption{Comparison of action head complexity and inference efficiency. Stochastic indicates whether the action head explicitly models the action distribution. Params denotes the number of parameters in the action head. Latency denotes the mean time required by the action head to decode one action chunk, excluding the VLA backbone. SURE$^\ast$ denotes our JAX variant used for a framework-matched comparison with $\pi_{0.5}$.}
    \label{tab:inference_efficiency}
    {
    \small
    \begin{tabular}{
        >{\centering\arraybackslash}m{1.4cm}
        >{\arraybackslash}m{2.7cm}
        >{\centering\arraybackslash}m{1.15cm}
        >{\centering\arraybackslash}m{1.0cm}
        >{\centering\arraybackslash}m{1.08cm}
        }
        \toprule
        Stochastic                  & Methods           & Params  & FLOPs  & Latency \\
        \midrule
        No                          & OFT               & 151.12M & 2.42G  & 0.37ms  \\
        \midrule
        \multirow{3}{*}{Yes}        & $\pi_{0.5}$ (JAX) & 430.10M & 79.13G & 19.88ms \\
                                    & SURE              & 16.93M  & 10.96G & 1.57ms  \\
                                    & SURE$^\ast$ (JAX)  & 16.93M  & 10.96G & 0.69ms  \\
            \bottomrule
        \end{tabular}
    }
\end{table}

\subsection{Data Efficiency Evaluation}

\noindent\textbf{Evaluation protocol.}
To evaluate data efficiency independently of the optimization budget, we train SURE and OpenVLA-OFT on LIBERO-Spatial using 10\%, 25\%, 50\%, and 100\% of the available training demonstrations. We fix the training budget at 10k optimization steps for all data ratios and keep the remaining training settings unchanged. OpenVLA-OFT serves as the baseline because it uses the same OpenVLA-7B backbone and provides a strong fine-tuning recipe for continuous action prediction \citep{kim2025oft}. All results in this evaluation are obtained using a single seed.

\begin{figure}[htb]
    \centering
    \includegraphics[width=0.5\linewidth]{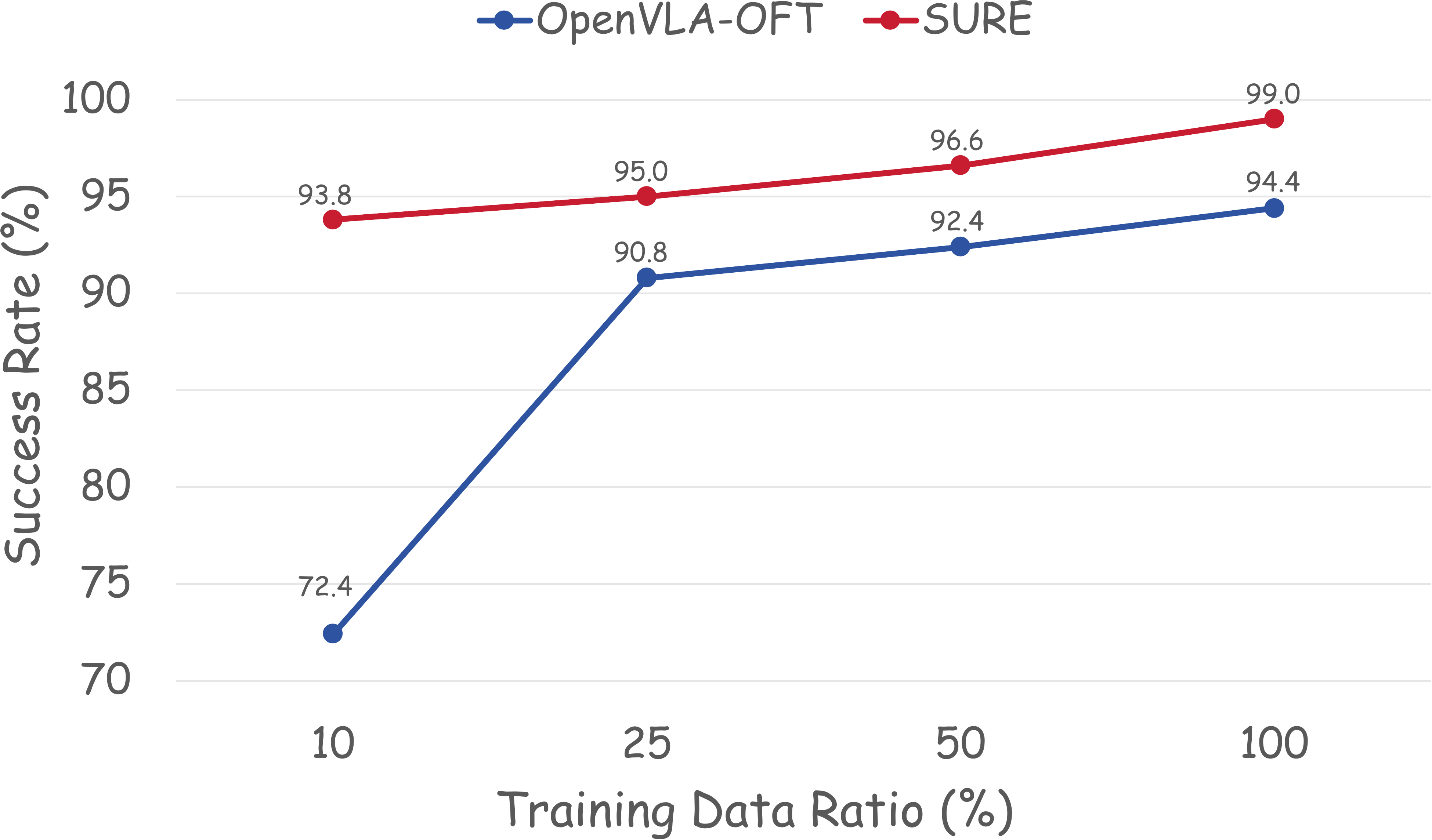}
    \caption{Success rates (\%) on LIBERO-Spatial using different fractions of the available training demonstrations. Both methods are trained for 10k optimization steps.}
    \label{fig:data_efficiency}
\end{figure}

\noindent\textbf{Results.}
As shown in Figure~\ref{fig:data_efficiency}, SURE consistently outperforms OpenVLA-OFT across all training data ratios. With 10\%, 25\%, 50\%, and 100\% of the demonstrations, SURE improves the success rate by 21.4, 4.2, 4.2, and 4.6\%, respectively. Notably, SURE reaches 95.0\% success using only 25\% of the training data, exceeding the 94.4\% achieved by OpenVLA-OFT with the full dataset. Even with only 10\% of the demonstrations, SURE achieves 93.8\%, remaining within 0.6\% of the full-data OpenVLA-OFT result. These results show that explicitly modeling spatial and temporal uncertainty allows SURE to learn effective manipulation policies from substantially fewer demonstrations.

\subsection{Ablations}

We conduct subtractive ablations on LIBERO-Spatial to evaluate the probabilistic action decoder and Soft-DTW alignment supervision. All variants are trained on LIBERO-Spatial for 10k optimization steps using the same settings.

As shown in Table~\ref{tab:ablation_modules}, removing either component reduces the success rate. Removing Soft-DTW decreases performance from 99.0\% to 97.4\%, while removing the probabilistic action decoder yields 96.8\%. Removing both components further reduces the success rate to 94.4\%, confirming that spatial and temporal uncertainty modeling are complementary.

\begin{table}[htb]
    \centering
    \caption{Subtractive ablation on LIBERO-Spatial. Prob denotes the probabilistic action decoder, and DTW denotes Soft-DTW alignment supervision. All variants are trained for 10k steps.}
    \label{tab:ablation_modules}
    {
        \small
        \begin{tabular}{
            >{\arraybackslash}m{2.0cm}
            >{\centering\arraybackslash}m{1.1cm}
            >{\centering\arraybackslash}m{0.9cm}
            >{\centering\arraybackslash}m{0.9cm}
            >{\centering\arraybackslash}m{2.0cm}
            }
            \toprule
            Variant      & Steps & Prob       & DTW        & Success Rate  \\
            \midrule
            \textbf{SURE} & 10k   & \checkmark & \checkmark & \textbf{99.0} \\
            w/o DTW      & 10k   & \checkmark & --         & 97.4          \\
            w/o Prob     & 10k   & --         & \checkmark & 96.8          \\
            w/o Both     & 10k   & --         & --         & 94.4          \\
            \bottomrule
        \end{tabular}
    }
\end{table}

\section{Conclusion}

We presented SURE, a training- and data-efficient VLA framework that explicitly models the spatial and temporal uncertainty inherent in robot demonstrations. SURE combines a lightweight probabilistic action decoder for spatial uncertainty with Soft-DTW alignment supervision that accommodates differences in execution timing. The decoder further supports fast parallel action prediction without iterative denoising. Experiments on LIBERO, RoboTwin 2.0, and six real-world tasks demonstrate improved task performance and data efficiency under reduced training budgets, together with efficient action decoding. Future work may extend SURE to larger cross-embodiment datasets and more diverse manipulation settings.

\subsection*{AI Use Statement}

In this work, we used generative AI tools for language editing, manuscript organization, \LaTeX{} formatting, and summarizing experiment records provided by the authors for presentation in the manuscript. We did not use generative AI tools to design the methodology, write or modify experimental code, run experiments, or generate experimental data or results. Other disclosure tasks are not applicable to this work. We reviewed all work produced with AI assistance. The authors checked text produced with AI assistance against the manuscript plan and original experiment records, verified citation metadata against primary sources, and compiled and visually inspected the affected pages. We take responsibility for the final content of this work, including text, claims, or artifacts produced with the aid of generative AI.

\subsection*{Ethics statement}

This work evaluates robot manipulation policies on the LIBERO and RoboTwin 2.0 simulation benchmarks and on six tabletop tasks using LeRobot SO-101 arms. The real-world demonstrations were collected by the authors as operators. The tasks involve robot interaction with objects rather than interaction with people. Physical deployment nevertheless carries risks from unintended arm motion, collisions, and spills during the pouring task. The reported task success rates do not measure operational safety or justify unsupervised use around people. Deployment beyond the evaluated settings would require task-specific safety assessment, appropriate physical safeguards, and human oversight.

\subsection*{Reproducibility statement}

The Methods section defines the probabilistic action decoder, Soft-DTW alignment, and training objective. Section~\ref{sec:experiments} specifies the LIBERO, RoboTwin 2.0, and real-world evaluation protocols, including task selection, training budgets, evaluation trials, and the provenance of baseline results. It also describes the data-efficiency, ablation, and action-head latency evaluations. Appendix~\ref{app:training-details} reports the backbone adaptation, optimizer and loss settings, action chunk lengths, training schedules, hardware, and the RoboTwin-specific decoder. Appendix~\ref{app:real-world} describes the robot setup, demonstration collection, six task instructions, and representative rollouts.

\bibliography{sure_preprint}

@inproceedings{brohan2022rt1,
  title     = {{RT-1}: Robotics Transformer for Real-World Control at Scale},
  author    = {Brohan, Anthony and Brown, Noah and Carbajal, Justice and Chebotar, Yevgen and Dabis, Joseph and Finn, Chelsea and Gopalakrishnan, Keerthana and Hausman, Karol and Herzog, Alexander and Hsu, Jasmine and Ibarz, Julian and Ichter, Brian and Irpan, Alex and Jackson, Tomas and Jesmonth, Sally and Joshi, Nikhil and Julian, Ryan and Kalashnikov, Dmitry and Kuang, Yuheng and Leal, Isabel and Lee, Kuang-Huei and Levine, Sergey and Lu, Yao and Malla, Utsav and Manjunath, Deeksha and Mordatch, Igor and Nachum, Ofir and Parada, Carolina and Peralta, Jodilyn and Perez, Emily and Pertsch, Karl and Quiambao, Jornell and Rao, Kanishka and Ryoo, Michael S. and Salazar, Grecia and Sanketi, Pannag R. and Sayed, Kevin and Singh, Jaspiar and Sontakke, Sumedh and Stone, Austin and Tan, Clayton and Tran, Huong and Vanhoucke, Vincent and Vega, Steve and H Vuong, Quan and Xia, Fei and Xiao, Ted and Xu, Peng and Xu, Sichun and Yu, Tianhe and Zitkovich, Brianna},
  booktitle = {Proceedings of Robotics: Science and Systems},
  year      = {2023},
  url       = {https://www.roboticsproceedings.org/rss19/p025.html},
  doi       = {10.15607/RSS.2023.XIX.025}
}

@inproceedings{ghosh2024octo,
  title     = {Octo: An Open-Source Generalist Robot Policy},
  author    = {Ghosh, Dibya and Walke, Homer Rich and Pertsch, Karl and Black, Kevin and Mees, Oier and Dasari, Sudeep and Hejna, Joey and Kreiman, Tobias and Xu, Charles and Luo, Jianlan and Tan, You Liang and Chen, Lawrence Yunliang and Vuong, Quan and Xiao, Ted and Sanketi, Pannag R. and Sadigh, Dorsa and Finn, Chelsea and Levine, Sergey},
  booktitle = {Proceedings of Robotics: Science and Systems},
  year      = {2024},
  url       = {https://www.roboticsproceedings.org/rss20/p090.html},
  doi       = {10.15607/RSS.2024.XX.090}
}

@inproceedings{oxe2023rtx,
  title     = {{Open X-Embodiment}: Robotic Learning Datasets and {RT-X} Models},
  author    = {{Open X-Embodiment Collaboration} and O'Neill, Abby and Rehman, Abdul and Maddukuri, Abhiram and Gupta, Abhishek and Padalkar, Abhishek and Lee, Abraham and Pooley, Acorn and Gupta, Agrim and Mandlekar, Ajay and Jain, Ajinkya and Tung, Albert and Bewley, Alex and Herzog, Alex and Irpan, Alex and Khazatsky, Alexander and Rai, Anant and Gupta, Anchit and Wang, Andrew and Singh, Anikait and Garg, Animesh and Kembhavi, Aniruddha and Xie, Annie and Brohan, Anthony and Raffin, Antonin and Sharma, Archit and Yavary, Arefeh and Jain, Arhan and Balakrishna, Ashwin and Wahid, Ayzaan and Burgess-Limerick, Ben and Kim, Beomjoon and Schölkopf, Bernhard and Wulfe, Blake and Ichter, Brian and Lu, Cewu and Xu, Charles and Le, Charlotte and Finn, Chelsea and Wang, Chen and Xu, Chenfeng and Chi, Cheng and Huang, Chenguang and Chan, Christine and Agia, Christopher and Pan, Chuer and Fu, Chuyuan and Devin, Coline and Xu, Danfei and Morton, Daniel and Driess, Danny and Chen, Daphne and Pathak, Deepak and Shah, Dhruv and Büchler, Dieter and Jayaraman, Dinesh and Kalashnikov, Dmitry and Sadigh, Dorsa and Johns, Edward and Foster, Ethan and Liu, Fangchen and Ceola, Federico and Xia, Fei and Zhao, Feiyu and Stulp, Freek and Zhou, Gaoyue and Sukhatme, Gaurav S. and Salhotra, Gautam and Yan, Ge and Feng, Gilbert and Schiavi, Giulio and Berseth, Glen and Kahn, Gregory and Wang, Guanzhi and Su, Hao and Fang, Hao-Shu and Shi, Haochen and Bao, Henghui and Ben Amor, Heni and Christensen, Henrik I and Furuta, Hiroki and Walke, Homer and Fang, Hongjie and Ha, Huy and Mordatch, Igor and Radosavovic, Ilija and Leal, Isabel and Liang, Jacky and Abou-Chakra, Jad and Kim, Jaehyung and Drake, Jaimyn and Peters, Jan and Schneider, Jan and Hsu, Jasmine and Bohg, Jeannette and Bingham, Jeffrey and Wu, Jeffrey and Gao, Jensen and Hu, Jiaheng and Wu, Jiajun and Wu, Jialin and Sun, Jiankai and Luo, Jianlan and Gu, Jiayuan and Tan, Jie and Oh, Jihoon and Wu, Jimmy and Lu, Jingpei and Yang, Jingyun and Malik, Jitendra and Silvério, João and Hejna, Joey and Booher, Jonathan and Tompson, Jonathan and Yang, Jonathan and Salvador, Jordi and Lim, Joseph J. and Han, Junhyek and Wang, Kaiyuan and Rao, Kanishka and Pertsch, Karl and Hausman, Karol and Go, Keegan and Gopalakrishnan, Keerthana and Goldberg, Ken and Byrne, Kendra and Oslund, Kenneth and Kawaharazuka, Kento and Black, Kevin and Lin, Kevin and Zhang, Kevin and Ehsani, Kiana and Lekkala, Kiran and Ellis, Kirsty and Rana, Krishan and Srinivasan, Krishnan and Fang, Kuan and Singh, Kunal Pratap and Zeng, Kuo-Hao and Hatch, Kyle and Hsu, Kyle and Itti, Laurent and Chen, Lawrence Yunliang and Pinto, Lerrel and Fei-Fei, Li and Tan, Liam and Fan, Linxi Jim and Ott, Lionel and Lee, Lisa and Weihs, Luca and Chen, Magnum and Lepert, Marion and Memmel, Marius and Tomizuka, Masayoshi and Itkina, Masha and Castro, Mateo Guaman and Spero, Max and Du, Maximilian and Ahn, Michael and Yip, Michael C. and Zhang, Mingtong and Ding, Mingyu and Heo, Minho and Srirama, Mohan Kumar and Sharma, Mohit and Kim, Moo Jin and Kanazawa, Naoaki and Hansen, Nicklas and Heess, Nicolas and Joshi, Nikhil J and Suenderhauf, Niko and Liu, Ning and Di Palo, Norman and Shafiullah, Nur Muhammad Mahi and Mees, Oier and Kroemer, Oliver and Bastani, Osbert and Sanketi, Pannag R and Miller, Patrick Tree and Yin, Patrick and Wohlhart, Paul and Xu, Peng and Fagan, Peter David and Mitrano, Peter and Sermanet, Pierre and Abbeel, Pieter and Sundaresan, Priya and Chen, Qiuyu and Vuong, Quan and Rafailov, Rafael and Tian, Ran and Doshi, Ria and Martín-Martín, Roberto and Baijal, Rohan and Scalise, Rosario and Hendrix, Rose and Lin, Roy and Qian, Runjia and Zhang, Ruohan and Mendonca, Russell and Shah, Rutav and Hoque, Ryan and Julian, Ryan and Bustamante, Samuel and Kirmani, Sean and Levine, Sergey and Lin, Shan and Moore, Sherry and Bahl, Shikhar and Dass, Shivin and Sonawani, Shubham and Song, Shuran and Xu, Sichun and Haldar, Siddhant and Karamcheti, Siddharth and Adebola, Simeon and Guist, Simon and Nasiriany, Soroush and Schaal, Stefan and Welker, Stefan and Tian, Stephen and Ramamoorthy, Subramanian and Dasari, Sudeep and Belkhale, Suneel and Park, Sungjae and Nair, Suraj and Mirchandani, Suvir and Osa, Takayuki and Gupta, Tanmay and Harada, Tatsuya and Matsushima, Tatsuya and Xiao, Ted and Kollar, Thomas and Yu, Tianhe and Ding, Tianli and Davchev, Todor and Zhao, Tony Z. and Armstrong, Travis and Darrell, Trevor and Chung, Trinity and Jain, Vidhi and Vanhoucke, Vincent and Zhan, Wei and Zhou, Wenxuan and Burgard, Wolfram and Chen, Xi and Wang, Xiaolong and Zhu, Xinghao and Geng, Xinyang and Liu, Xiyuan and Liangwei, Xu and Li, Xuanlin and Lu, Yao and Ma, Yecheng Jason and Kim, Yejin and Chebotar, Yevgen and Zhou, Yifan and Zhu, Yifeng and Wu, Yilin and Xu, Ying and Wang, Yixuan and Bisk, Yonatan and Cho, Yoonyoung and Lee, Youngwoon and Cui, Yuchen and Cao, Yue and Wu, Yueh-Hua and Tang, Yujin and Zhu, Yuke and Zhang, Yunchu and Jiang, Yunfan and Li, Yunshuang and Li, Yunzhu and Iwasawa, Yusuke and Matsuo, Yutaka and Ma, Zehan and Xu, Zhuo and Cui, Zichen Jeff and Zhang, Zichen and Lin, Zipeng},
  booktitle = {Proceedings of International Conference on Robotics \& Automation},
  year      = {2024},
  url       = {https://ieeexplore.ieee.org/document/10611477},
  doi       = {10.1109/ICRA57147.2024.10611477},
  pages     = {6892--6903}
}

@inproceedings{kim2024openvla,
  title     = {{OpenVLA}: An Open-Source Vision-Language-Action Model},
  author    = {Kim, Moo Jin and Pertsch, Karl and Karamcheti, Siddharth and Xiao, Ted and Balakrishna, Ashwin and Nair, Suraj and Rafailov, Rafael and Foster, Ethan P and Sanketi, Pannag R and Vuong, Quan and Kollar, Thomas and Burchfiel, Benjamin and Tedrake, Russ and Sadigh, Dorsa and Levine, Sergey and Liang, Percy and Finn, Chelsea},
  booktitle = {Proceedings of Conference on Robot Learning},
  year      = {2025},
  url       = {https://proceedings.mlr.press/v270/kim25c.html},
  pages     = {2679--2713}
}

@inproceedings{Black2024Pi0,
  title     = {{$\pi_0$}: A Vision-Language-Action Flow Model for General Robot Control},
  author    = {Black, Kevin and Brown, Noah and Driess, Danny and Esmail, Adnan and Equi, Michael Robert and Finn, Chelsea and Fusai, Niccolo and Groom, Lachy and Hausman, Karol and Ichter, Brian and Jakubczak, Szymon and Jones, Tim and Ke, Liyiming and Levine, Sergey and Li-Bell, Adrian and Mothukuri, Mohith and Nair, Suraj and Pertsch, Karl and Shi, Lucy Xiaoyang and Smith, Laura and Tanner, James and Vuong, Quan and Walling, Anna and Wang, Haohuan and Zhilinsky, Ury},
  year      = {2025},
  booktitle = {Proceedings of Robotics: Science and Systems},
  url       = {https://www.roboticsproceedings.org/rss21/p010.html},
  doi       = {10.15607/RSS.2025.XXI.010}
}

@inproceedings{chi2023diffusionpolicy,
  title     = {{Diffusion Policy}: Visuomotor Policy Learning via Action Diffusion},
  author    = {Chi, Cheng and Feng, Siyuan and Du, Yilun and Xu, Zhenjia and Cousineau, Eric and CM Burchfiel, Benjamin and Song, Shuran},
  booktitle = {Proceedings of Robotics: Science and Systems},
  year      = {2023},
  url       = {https://www.roboticsproceedings.org/rss19/p026.html},
  doi       = {10.15607/RSS.2023.XIX.026}
}

@article{bjorck2025gr00t,
  title   = {{GR00T N1}: An Open Foundation Model for Generalist Humanoid Robots},
  author  = {{NVIDIA} and Bjorck, Johan and Castañeda, Fernando and Cherniadev, Nikita and Da, Xingye and Ding, Runyu and Fan, Linxi Jim and Fang, Yu and Fox, Dieter and Hu, Fengyuan and Huang, Spencer and Jang, Joel and Jiang, Zhenyu and Kautz, Jan and Kundalia, Kaushil and Lao, Lawrence and Li, Zhiqi and Lin, Zongyu and Lin, Kevin and Liu, Guilin and Llontop, Edith and Magne, Loic and Mandlekar, Ajay and Narayan, Avnish and Nasiriany, Soroush and Reed, Scott and Tan, You Liang and Wang, Guanzhi and Wang, Zu and Wang, Jing and Wang, Qi and Xiang, Jiannan and Xie, Yuqi and Xu, Yinzhen and Xu, Zhenjia and Ye, Seonghyeon and Yu, Zhiding and Zhang, Ao and Zhang, Hao and Zhao, Yizhou and Zheng, Ruijie and Zhu, Yuke},
  journal = {CoRR},
  year    = {2025},
  url     = {https://arxiv.org/abs/2503.14734},
  eprint  = {2503.14734}
}

@inproceedings{hafner2019dreamer,
  title     = {Dream to Control: Learning Behaviors by Latent Imagination},
  author    = {Hafner, Danijar and Lillicrap, Timothy and Ba, Jimmy and Norouzi, Mohammad},
  booktitle = {Proceedings of International Conference on Learning Representations},
  year      = {2020},
  url       = {https://arxiv.org/abs/1912.01603}
}

@inproceedings{liao2025genie,
  title     = {{Genie Envisioner}: A Unified World Foundation Platform for Robotic Manipulation},
  author    = {Liao, Yue and Zhou, Pengfei and Huang, Siyuan and Yang, Donglin and Chen, Shengcong and Jiang, Yuxin and Hu, Yue and Liu, Si and Luo, Jianlan and Chen, Liliang and YAN, Shuicheng and Yao, Maoqing and Ren, Guanghui},
  booktitle = {Proceedings of International Conference on Learning Representations},
  year      = {2026},
  url       = {https://proceedings.iclr.cc/paper_files/paper/2026/hash/8e773e9e463e97b930c47621b6777159-Abstract-Conference.html}
}

@inproceedings{guo2025ctrlworld,
  title     = {{Ctrl-World}: A Controllable Generative World Model for Robot Manipulation},
  author    = {Guo, Yanjiang and Shi, Lucy and Chen, Jianyu and Finn, Chelsea},
  booktitle = {Proceedings of International Conference on Learning Representations},
  pages     = {6121--6138},
  year      = {2026},
  url       = {https://proceedings.iclr.cc/paper_files/paper/2026/hash/0ae94013da7cd459402fd77874e09ee3-Abstract-Conference.html}
}

@article{zhou2026tau0wm,
  title   = {{$\tau_0$}-{WM}: A Unified Video-Action World Model for Robotic Manipulation},
  author  = {Zhou, Pengfei and Chen, Shengcong and Chen, Di and Wang, Jiaxu and Jin, Rongjun and Zhu, Bingwen and Pan, Yike and Gu, Songen and Wang, Kuanning and Nan, Shufeng and Qiu, Xingyu and Qiu, Chenhao and Yang, Pu and Cai, Yunuo and Gao, Jianxiong and Li, Yifan and Fu, Yanwei and Yue, Xiangyu and Chen, Zhi and Luo, Jianlan},
  journal = {CoRR},
  year    = {2026},
  url     = {https://arxiv.org/abs/2606.01027},
  eprint  = {2606.01027}
}

@article{sakoe1978dtw,
  title   = {Dynamic Programming Algorithm Optimization for Spoken Word Recognition},
  author  = {Sakoe, Hiroaki and Chiba, Seibi},
  journal = {IEEE Transactions on Acoustics, Speech, and Signal Processing},
  volume  = {26},
  number  = {1},
  pages   = {43--49},
  year    = {1978},
  doi     = {10.1109/TASSP.1978.1163055},
  url     = {https://doi.org/10.1109/TASSP.1978.1163055}
}

@inproceedings{cuturi2017softdtw,
  title     = {{Soft-DTW}: a Differentiable Loss Function for Time-Series},
  author    = {Cuturi, Marco and Blondel, Mathieu},
  booktitle = {Proceedings of International Conference on Machine Learning},
  pages     = {894--903},
  year      = {2017},
  url       = {https://proceedings.mlr.press/v70/cuturi17a.html}
}

@inproceedings{pertsch2025fast,
  title     = {{FAST}: Efficient Action Tokenization for Vision-Language-Action Models},
  author    = {Pertsch, Karl and Stachowicz, Kyle and Ichter, Brian and Driess, Danny and Nair, Suraj and Vuong, Quan and Mees, Oier and Finn, Chelsea and Levine, Sergey},
  year      = {2025},
  booktitle = {Proceedings of Robotics: Science and Systems},
  url       = {https://www.roboticsproceedings.org/rss21/p012.html},
  doi       = {10.15607/RSS.2025.XXI.012}
}

@inproceedings{kim2025oft,
  title     = {Fine-Tuning Vision-Language-Action Models: Optimizing Speed and Success},
  author    = {Kim, Moo Jin and Finn, Chelsea and Liang, Percy},
  year      = {2025},
  booktitle = {Proceedings of Robotics: Science and Systems},
  url       = {https://www.roboticsproceedings.org/rss21/p017.html},
  doi       = {10.15607/RSS.2025.XXI.017}
}

@article{wen2024tinyvla,
  title   = {{TinyVLA}: Toward Fast, Data-Efficient Vision-Language-Action Models for Robotic Manipulation},
  author  = {Wen, Junjie and Zhu, Yichen and Li, Jinming and Zhu, Minjie and Tang, Zhibin and Wu, Kun and Xu, Zhiyuan and Liu, Ning and Cheng, Ran and Shen, Chaomin and Peng, Yaxin and Feng, Feifei and Tang, Jian},
  journal = {IEEE Robotics and Automation Letters},
  year    = {2025},
  url     = {https://ieeexplore.ieee.org/document/10900471/},
  doi     = {10.1109/LRA.2025.3544909}
}

@inproceedings{james2021c2f,
  title     = {{Coarse-to-Fine Q-attention}: Efficient Learning for Visual Robotic Manipulation via Discretisation},
  author    = {James, Stephen and Wada, Kentaro and Laidlow, Tristan and Davison, Andrew J.},
  booktitle = {Proceedings of IEEE/CVF Conference on Computer Vision and Pattern Recognition},
  year      = {2022},
  url       = {https://openaccess.thecvf.com/content/CVPR2022/html/James_Coarse-To-Fine_Q-Attention_Efficient_Learning_for_Visual_Robotic_Manipulation_via_Discretisation_CVPR_2022_paper.html},
  pages     = {13739--13748}
}

@inproceedings{shridhar2022peract,
  title     = {{Perceiver-Actor}: A Multi-Task Transformer for Robotic Manipulation},
  author    = {Shridhar, Mohit and Manuelli, Lucas and Fox, Dieter},
  booktitle = {Proceedings of Conference on Robot Learning},
  year      = {2023},
  url       = {https://proceedings.mlr.press/v205/shridhar23a.html},
  pages     = {785--799}
}

@inproceedings{goyal2023rvt,
  title     = {{RVT}: Robotic View Transformer for {3D} Object Manipulation},
  author    = {Goyal, Ankit and Xu, Jie and Guo, Yijie and Blukis, Valts and Chao, Yu-Wei and Fox, Dieter},
  booktitle = {Proceedings of Conference on Robot Learning},
  year      = {2023},
  url       = {https://proceedings.mlr.press/v229/goyal23a.html},
  pages     = {694--710}
}

@article{yang2026actionmap,
  title   = {{ActionMap}: Robot Policy Learning via Voxel Action Heatmap},
  author  = {Yang, Pei and Ci, Hai and Chen, Yanzhe and Lv, Qi and Cai, Han and Shou, Mike Zheng},
  journal = {CoRR},
  year    = {2026},
  url     = {https://arxiv.org/abs/2606.06904},
  eprint  = {2606.06904}
}

@inproceedings{liu2023libero,
  title     = {{LIBERO}: Benchmarking Knowledge Transfer for Lifelong Robot Learning},
  author    = {Liu, Bo and Zhu, Yifeng and Gao, Chongkai and Feng, Yihao and Liu, Qiang and Zhu, Yuke and Stone, Peter},
  booktitle = {Advances in Neural Information Processing Systems},
  year      = {2023},
  url       = {https://proceedings.neurips.cc/paper_files/paper/2023/hash/8c3c666820ea055a77726d66fc7d447f-Abstract-Datasets_and_Benchmarks.html}
}

@inproceedings{chen2025robotwin2,
  title     = {{RoboTwin} 2.0: A Scalable Data Generator and Benchmark with Strong Domain Randomization for Robust Bimanual Robotic Manipulation},
  author    = {Chen, Tianxing and Chen, Zanxin and Chen, Baijun and Cai, Zijian and Liu, Yibin and Li, Zixuan and Liang, Qiwei and Lin, Xianliang and Ge, Yiheng and Gu, Zhenyu and Deng, Weiliang and Guo, Yubin and Nian, Tian and Xie, Xuanbing and Chen, Qiangyu and Su, Kailun and Xu, Tianling and Liu, Guodong and Hu, Mengkang and Gao, Huan-ang and Wang, Kaixuan and Liang, Zhixuan and Qin, Yusen and Yang, Xiaokang and Luo, Ping and Mu, Yao},
  booktitle = {Proceedings of International Conference on Machine Learning},
  year      = {2026},
  url       = {https://icml.cc/virtual/2026/poster/62192}
}

@inproceedings{yeh2025acil,
  title     = {Action-Constrained Imitation Learning},
  author    = {Yeh, Chia-Han and Nan, Tse-Sheng and Vuorio, Risto and Hung, Wei and Wu, Hung-Yen and Sun, Shao-Hua and Hsieh, Ping-Chun},
  booktitle = {Proceedings of International Conference on Machine Learning},
  year      = {2025},
  url       = {https://proceedings.mlr.press/v267/yeh25a.html}
}

@article{song2026hume,
  title   = {Hume: Introducing System-2 Thinking in Visual-Language-Action Model},
  author  = {Song, Haoming and Qu, Delin and Yao, Yuanqi and Chen, Qizhi and Lv, Qi and Tang, Yiwen and Shi, Modi and Ren, Guanghui and Yao, Maoqing and Zhao, Bin and Wang, Dong and Li, Xuelong},
  journal = {CoRR},
  year    = {2025},
  url     = {https://arxiv.org/abs/2505.21432}
}

@article{miao2026jepavla,
  title   = {{JEPA-VLA}: Video Predictive Embedding is Needed for {VLA} Models},
  author  = {Shangchen Miao and Ningya Feng and Jialong Wu and Ye Lin and Xu He and Dong Li and Mingsheng Long},
  journal = {CoRR},
  year    = {2026},
  url     = {https://arxiv.org/abs/2602.11832}
}

@inproceedings{liu2025rdt,
  title     = {{RDT-1B}: a Diffusion Foundation Model for Bimanual Manipulation},
  author    = {Liu, Songming and Wu, Lingxuan and Li, Bangguo and Tan, Hengkai and Chen, Huayu and Wang, Zhengyi and Xu, Ke and Su, Hang and Zhu, Jun},
  booktitle = {Proceedings of International Conference on Learning Representations},
  year      = {2025},
  url       = {https://proceedings.iclr.cc/paper_files/paper/2025/hash/49f80e4d2471ad4f2edf4f5f1ab62339-Abstract-Conference.html}
}

@article{intelligence2025pi05,
  title   = {{$\pi_{0.5}$}: a Vision-Language-Action Model with Open-World Generalization},
  author  = {{Physical Intelligence} and Black, Kevin and Brown, Noah and Darpinian, James and Dhabalia, Karan and Driess, Danny and Esmail, Adnan and Equi, Michael and Finn, Chelsea and Fusai, Niccolo and Galliker, Manuel Y. and Ghosh, Dibya and Groom, Lachy and Hausman, Karol and Ichter, Brian and Jakubczak, Szymon and Jones, Tim and Ke, Liyiming and LeBlanc, Devin and Levine, Sergey and Li-Bell, Adrian and Mothukuri, Mohith and Nair, Suraj and Pertsch, Karl and Ren, Allen Z. and Shi, Lucy Xiaoyang and Smith, Laura and Springenberg, Jost Tobias and Stachowicz, Kyle and Tanner, James and Vuong, Quan and Walke, Homer and Walling, Anna and Wang, Haohuan and Yu, Lili and Zhilinsky, Ury},
  year    = {2025},
  journal = {CoRR},
  url     = {https://arxiv.org/abs/2504.16054}
}
\bibliographystyle{sure_preprint}

\clearpage
\appendix
\section{Implementation Details}\label{app:training-details}

Across all experiments, we adapt OpenVLA-7B using LoRA with rank 32 and $\alpha=16$, and optimize the model with AdamW using a weight decay of 0.01. We use an action chunk length of $k=8$ for LIBERO and the real-world experiments while $k=25$ for RoboTwin 2.0. The loss weights are $\lambda_{\mathrm{KL}}=0.05$ and $\lambda_{\mathrm{DTW}}=0.5$, and the Gaussian target distribution uses $\sigma=1.5$ bins.

\noindent\textbf{LIBERO.} We train one checkpoint for each task suite for 10k optimization steps on 4 NVIDIA H20 GPUs. We use a per-device batch size of 12, a learning rate of $5\times10^{-4}$, and an action chunk length of $k=8$. The Soft-DTW smoothing coefficient $\gamma$ is initialized to 0.1 and annealed to 0.005 over the first 5k optimization steps.

\noindent\textbf{RoboTwin 2.0.} Consistent with the baseline protocols, we train a separate checkpoint for each task. Each model is trained for 20k optimization steps on a single NVIDIA B300 GPU with a batch size of 16 and an action chunk length of $k=25$. The learning rate is initialized to $5\times10^{-4}$ and annealed to $5\times10^{-5}$ starting at step 10k. We initialize $\gamma$ to 0.1 and anneal it to 0.001 over the first 10k optimization steps.

LIBERO represents an action using end-effector translation, rotation, and a gripper command, whereas RoboTwin controls two arms using absolute joint angles. We therefore replace the translation--rotation decomposition in the probabilistic action decoder with a single 12-axis joint decoder, which predicts all 12 joint commands (six per arm) at each action step. Each joint is represented by a 256-bin distribution and decoded by expectation. A separate MLP branch predicts the two gripper commands, after which the output is rearranged to match RoboTwin's action convention, $[\mathbf{q}^{L}_{1:6}, g^{L}, \mathbf{q}^{R}_{1:6}, g^{R}]$. The stabilizing KL constraint is correspondingly applied to all 12 joint distributions. For the temporal objective, we first regroup each action as $[\mathbf{q}^{L}_{1:6}, \mathbf{q}^{R}_{1:6}, g^{L}, g^{R}]$ and apply Soft-DTW divergence to the predicted and ground-truth joint-action chunks.

\noindent\textbf{Real-world experiments.} We conduct real-world experiments using LeRobot SO-101 robotic arms. We train for 20k optimization steps on 2 NVIDIA A100 GPUs with a per-device batch size of 8, a learning rate of $5\times10^{-4}$, and an action chunk length of $k=8$. Following the same Soft-DTW schedule as RoboTwin 2.0, we anneal $\gamma$ from 0.1 to 0.001 over the first 10k optimization steps.

\section{Action Distributions Visualizations}

\subsection{LIBERO}

\begin{figure*}[htb]
    \centering
    \includegraphics[width=1.0\textwidth]{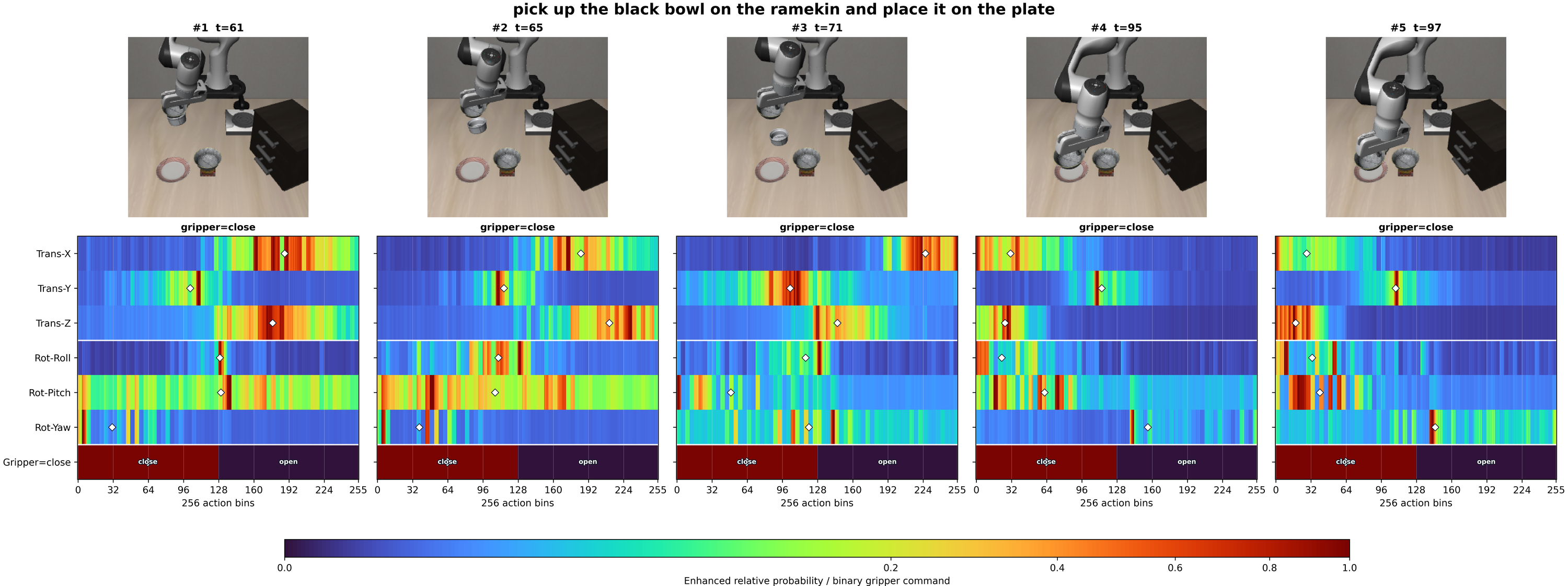}
    \includegraphics[width=1.0\textwidth]{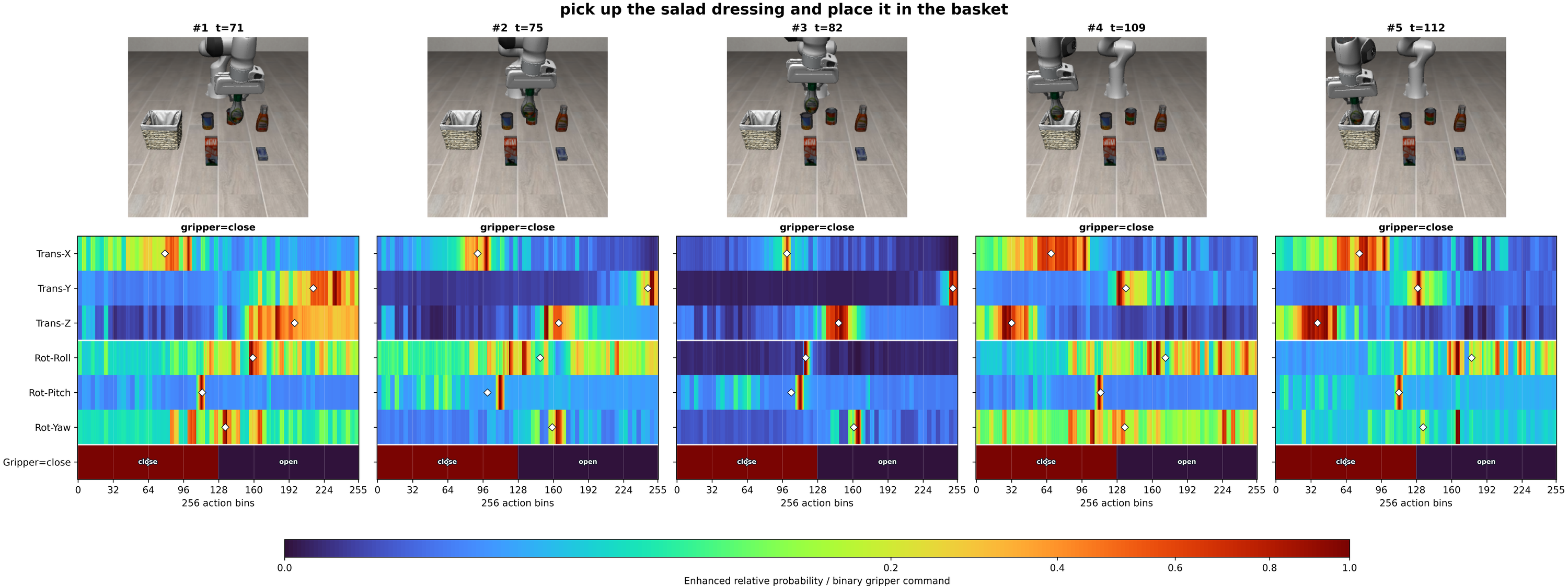}
    \includegraphics[width=1.0\textwidth]{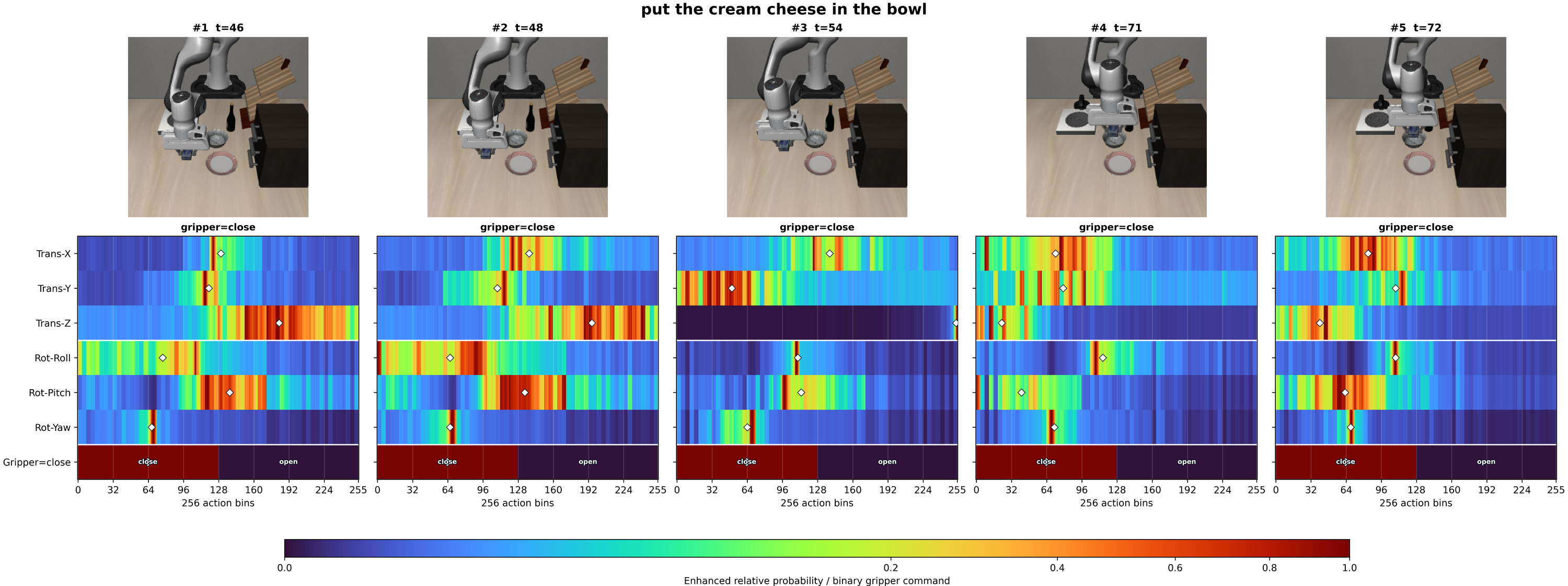}
    \includegraphics[width=1.0\textwidth]{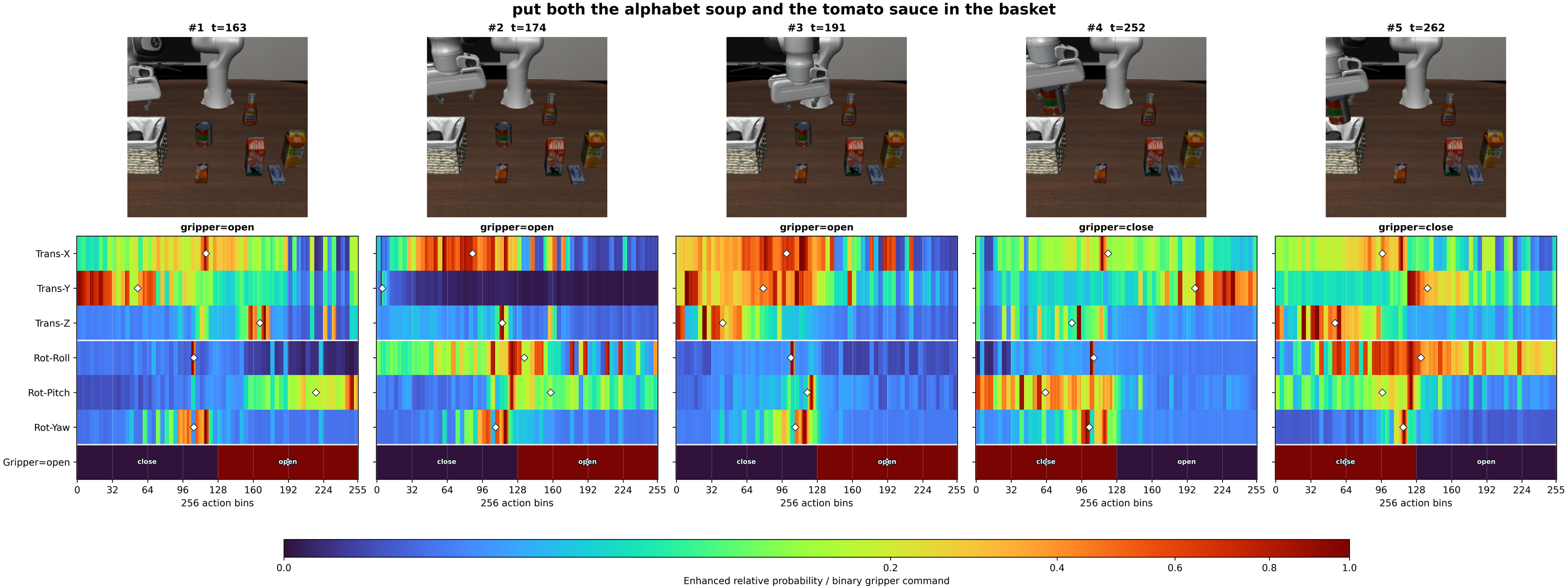}
    \caption{Representative SURE action-distribution visualizations on LIBERO-Spatial, LIBERO-Object, LIBERO-Goal, and LIBERO-Long, from top to bottom.}
    \label{fig:libero_visualizations}
\end{figure*}

Figure~\ref{fig:libero_visualizations} visualizes representative SURE action distributions on the four LIBERO task suites. In each column, the top image shows the robot observation and the bottom heatmaps show the predicted distributions for translation, rotation, and gripper actions. Diamonds mark the actions obtained by expectation-based decoding.

\subsection{RoboTwin 2.0}

\begin{figure*}[htb]
    \centering
    \includegraphics[width=1.0\textwidth]{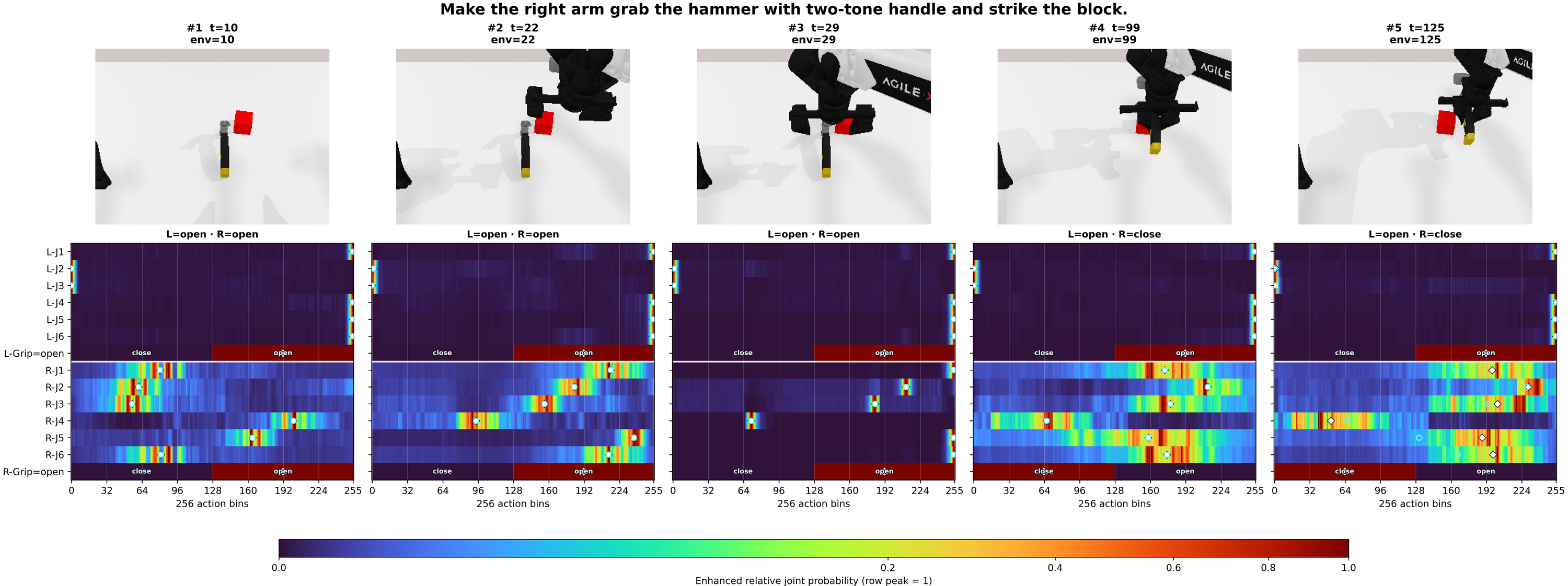}
    \includegraphics[width=1.0\textwidth]{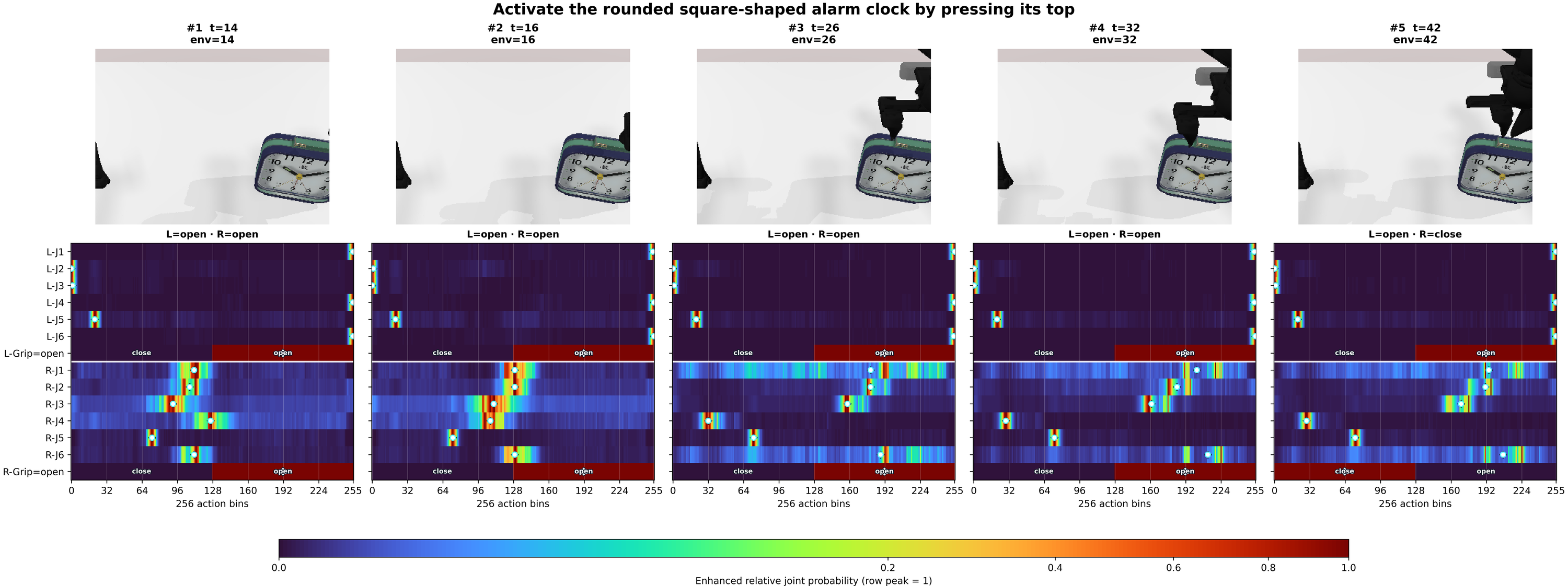}
    \includegraphics[width=1.0\textwidth]{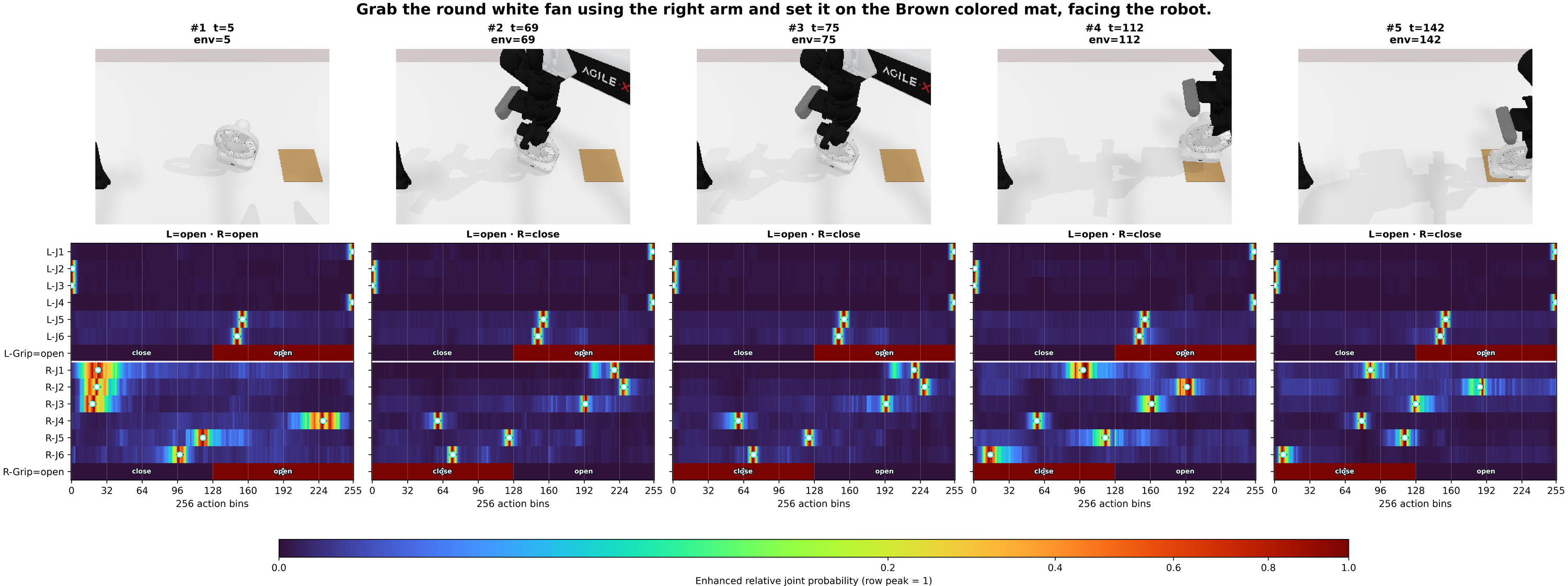}
    \includegraphics[width=1.0\textwidth]{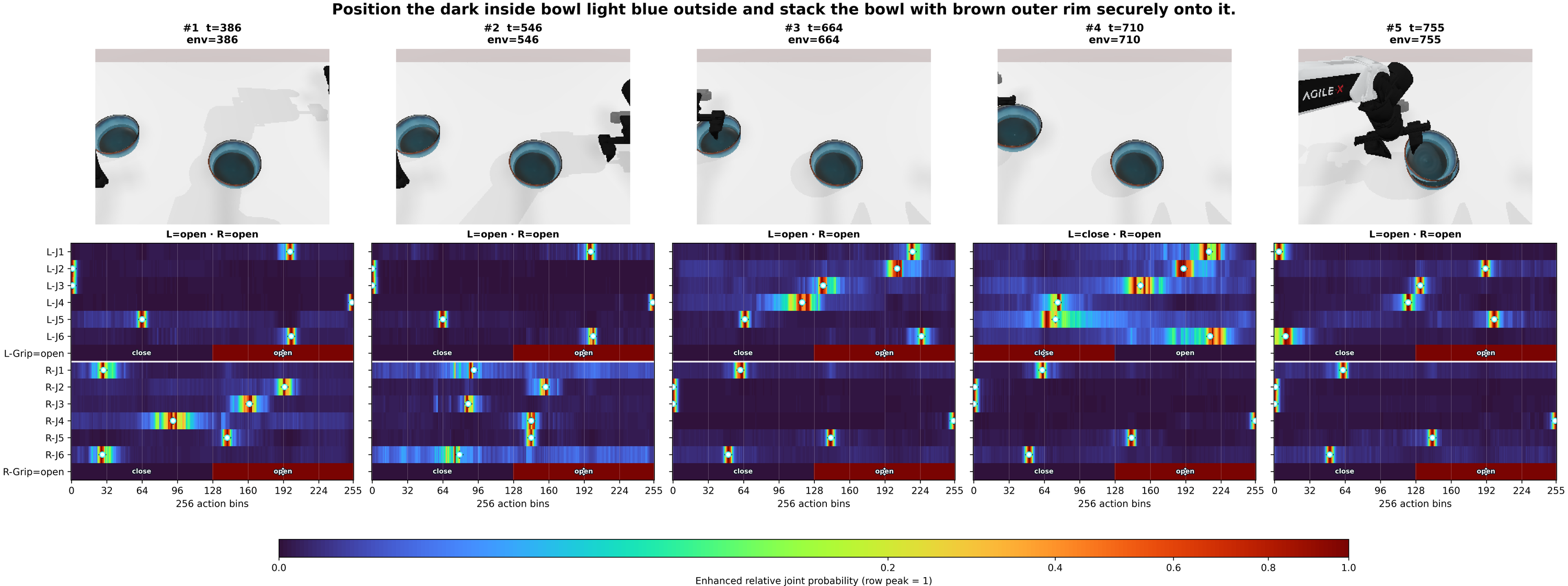}
    \caption{Representative SURE joint-action distribution visualizations on RoboTwin 2.0: Beat Block Hammer, Click Alarm Clock, Place Fan, and Stack Bowls Two, from top to bottom.}
    \label{fig:robotwin_visualizations}
\end{figure*}

Figure~\ref{fig:robotwin_visualizations} visualizes representative SURE joint-action distributions on four RoboTwin 2.0 tasks. In each column, the top image shows the bimanual robot observation, and the heatmaps below show the predicted action distributions for the left and right arm joints and grippers.

\newcommand{\taskrowlabel}[1]{\raisebox{0.28in}{\scriptsize\shortstack{#1}}}

\section{Real-World Experiments}\label{app:real-world}

\subsection{Setup Overview}

\begin{figure*}[htb]
    \centering
    \includegraphics[width=0.6\textwidth]{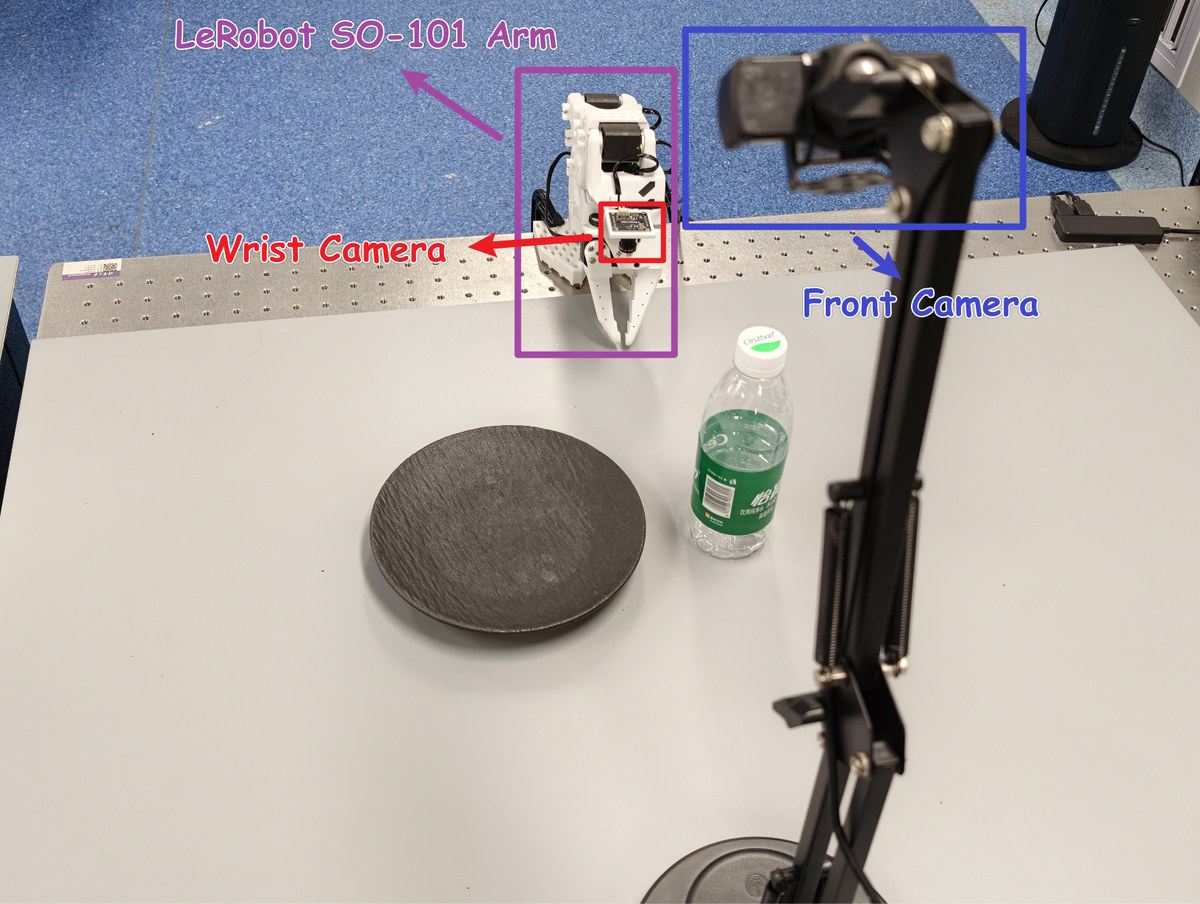}
    \caption{Real-world experimental setup with LeRobot SO-101 robotic arms. A front camera observes the workspace, while a wrist-mounted camera captures fine-grained manipulation details.}
    \label{fig:real_world_setup}
\end{figure*}

We conduct real-world experiments with LeRobot SO-101 robotic arms. As shown in Figure~\ref{fig:real_world_setup}, the setup provides a front camera and a wrist-mounted camera, which together capture the workspace and fine-grained manipulation details. We collect 50 demonstration episodes for each of the six tasks with multiple operators. The resulting variation in motion trajectories and execution speed provides natural spatial and temporal uncertainty. Both SURE and OpenVLA-OFT are jointly trained on the mixed-task demonstrations and evaluated using a single checkpoint. Each policy is evaluated over 20 trials per task, and we report the corresponding success rate.

\subsection{Instructions}

\begin{enumerate}
    \item \textbf{Water Bottle:} Pick up the bottle and place it on the black plate.
    \item \textbf{AAA Battery:} Pick up the AAA battery and place it in the white basket.
    \item \textbf{Screwdriver:} Pick up the screwdriver and place it in the white basket.
    \item \textbf{Toy Duck:} Pick up the toy duck and place it in the white basket.
    \item \textbf{Stack Bowls Two:} Stack the red bowl onto the black bowl.
    \item \textbf{Pour Water:} Pour the water from the water bottle into the cup.
\end{enumerate}

\subsection{Visualizations}

\begin{figure*}[htb]
    \centering
    \setlength{\tabcolsep}{1pt}
    \renewcommand{\arraystretch}{0.8}
    \begin{tabular}{@{}c*{5}{c}@{}}
        & {\scriptsize $t_0$} & {\scriptsize $t_1$} & {\scriptsize $t_2$} & {\scriptsize $t_3$} & {\scriptsize $t_4$} \\
        \taskrowlabel{Water\\Bottle}
        & \includegraphics[width=0.17\textwidth]{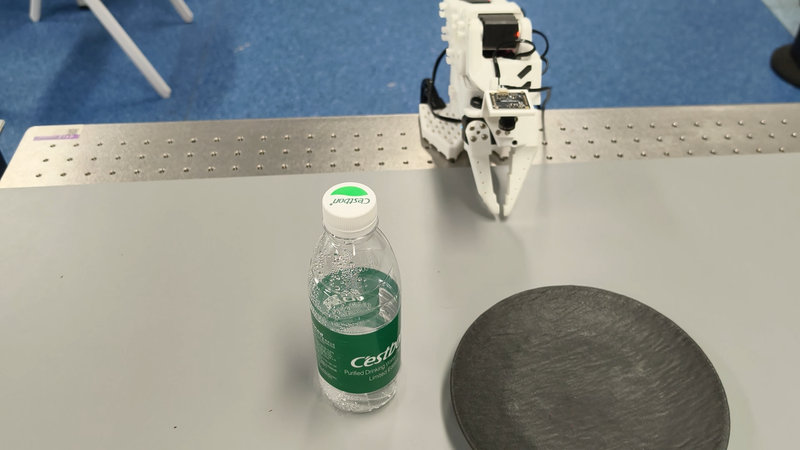}
        & \includegraphics[width=0.17\textwidth]{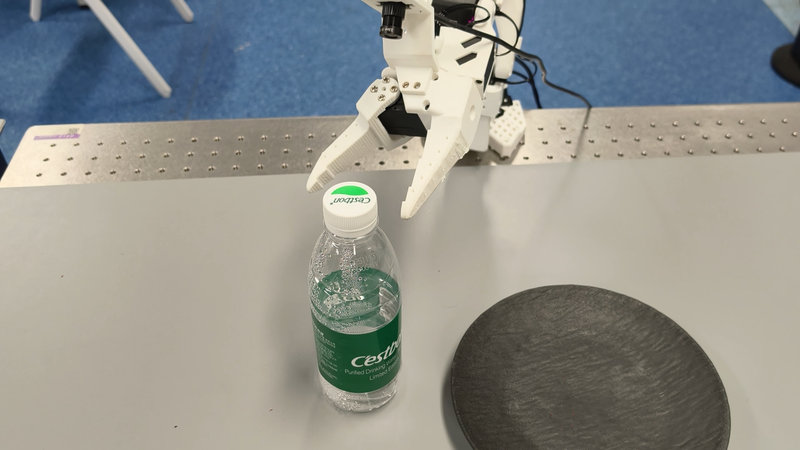}
        & \includegraphics[width=0.17\textwidth]{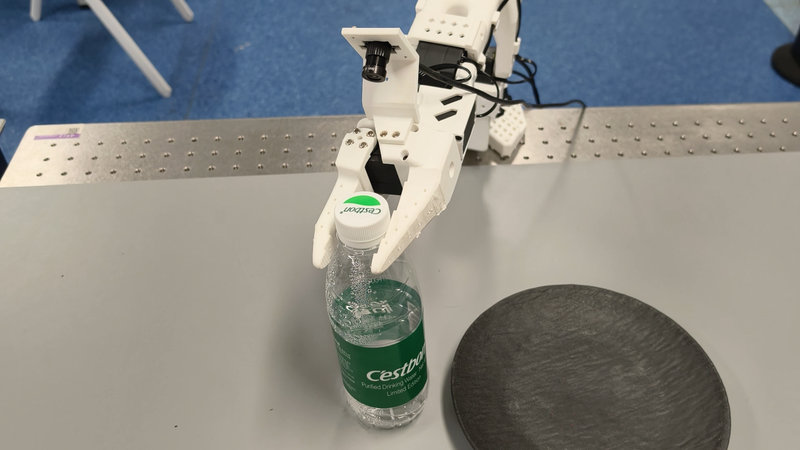}
        & \includegraphics[width=0.17\textwidth]{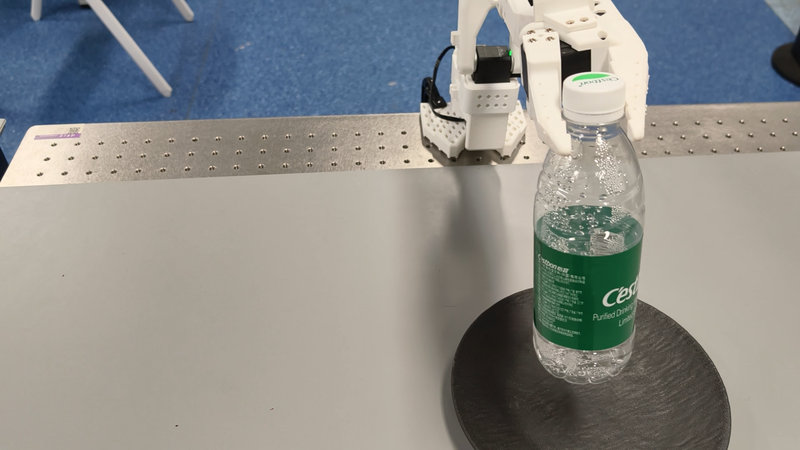}
        & \includegraphics[width=0.17\textwidth]{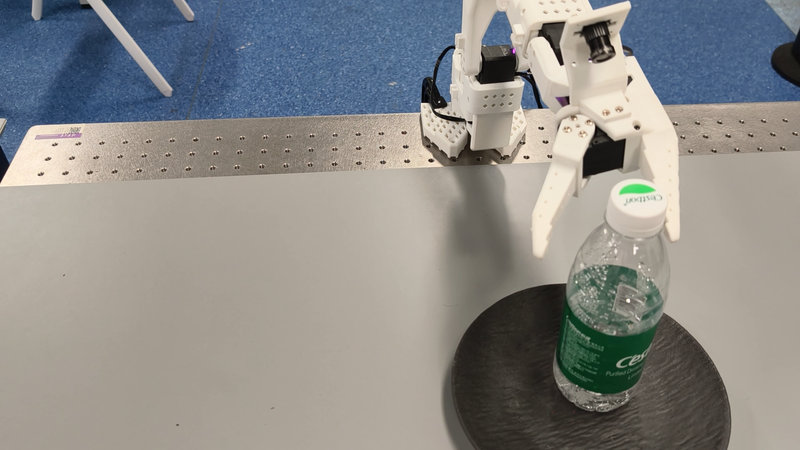} \\
        \taskrowlabel{AAA\\Battery}
        & \includegraphics[width=0.17\textwidth]{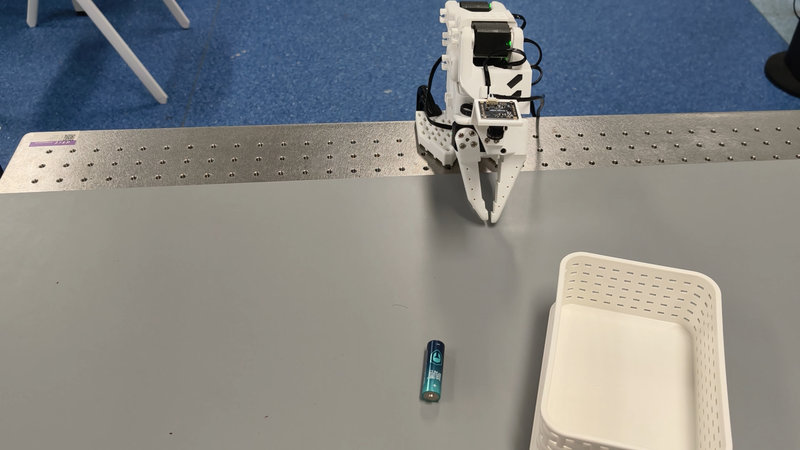}
        & \includegraphics[width=0.17\textwidth]{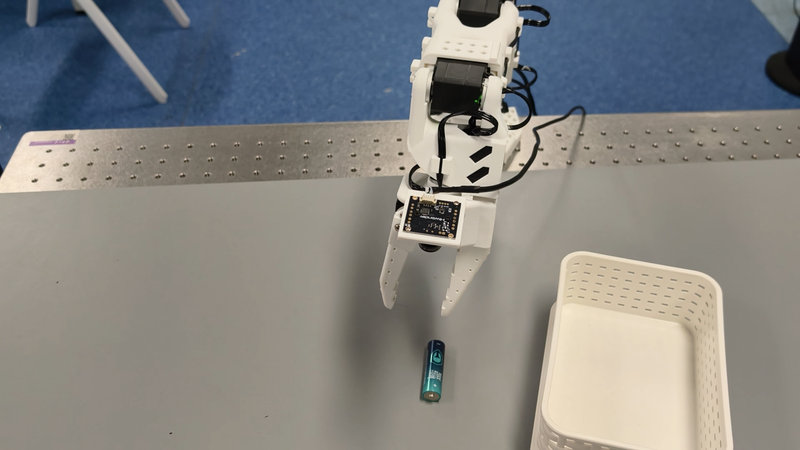}
        & \includegraphics[width=0.17\textwidth]{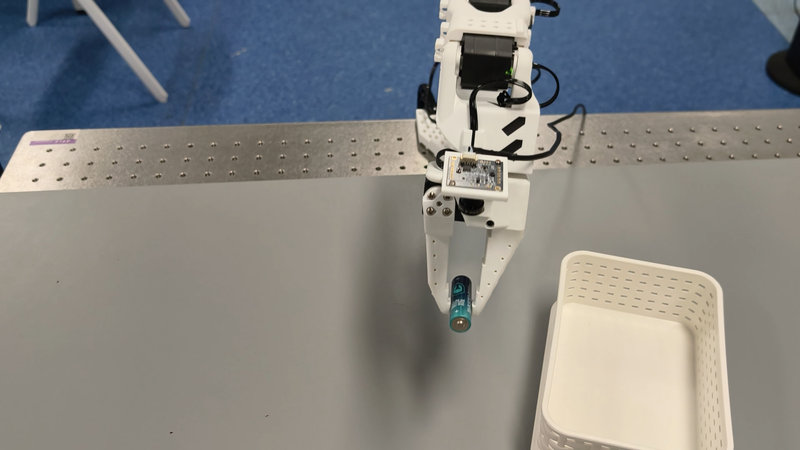}
        & \includegraphics[width=0.17\textwidth]{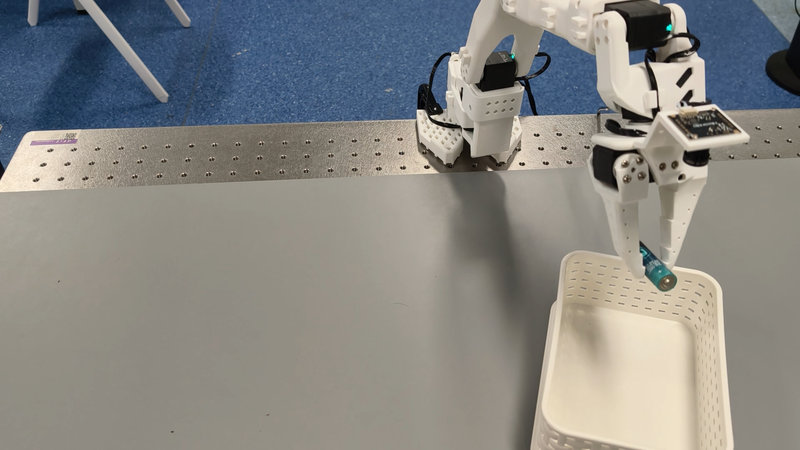}
        & \includegraphics[width=0.17\textwidth]{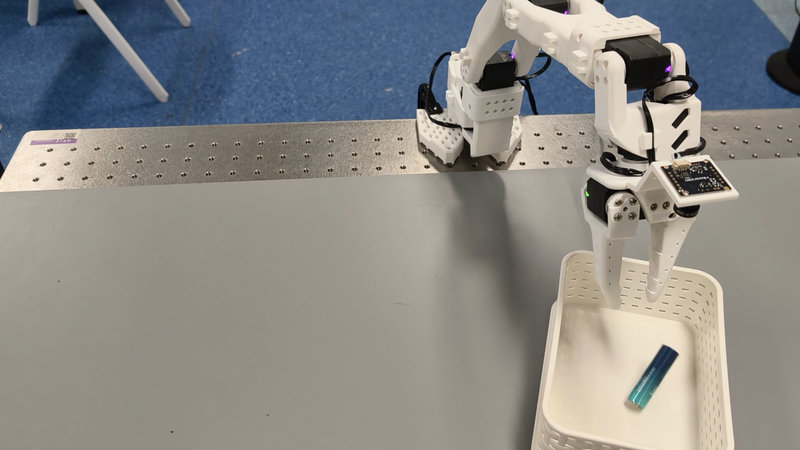} \\
        \taskrowlabel{Screwdriver}
        & \includegraphics[width=0.17\textwidth]{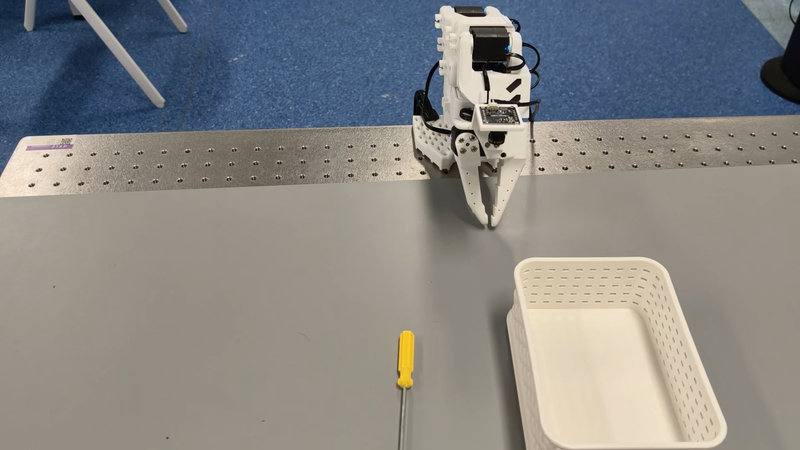}
        & \includegraphics[width=0.17\textwidth]{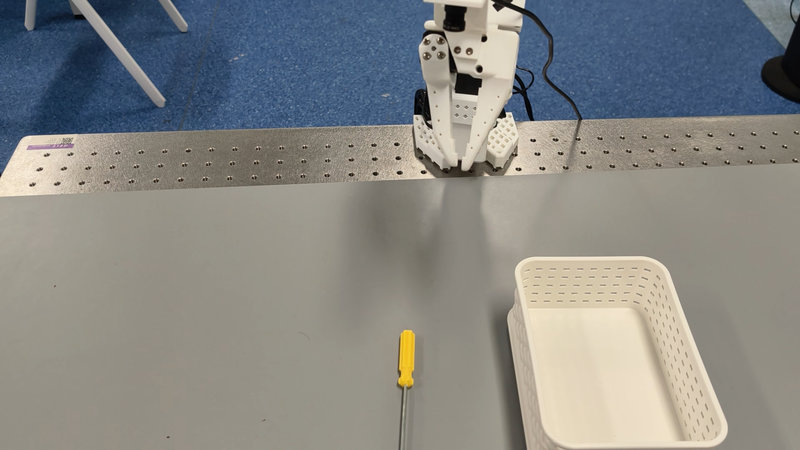}
        & \includegraphics[width=0.17\textwidth]{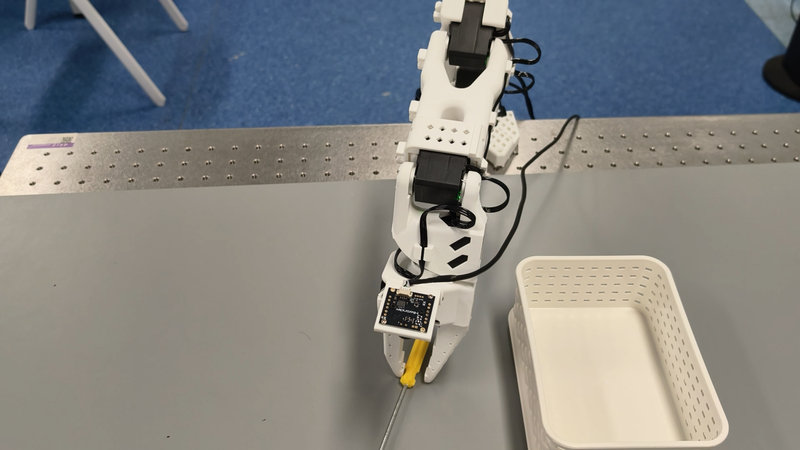}
        & \includegraphics[width=0.17\textwidth]{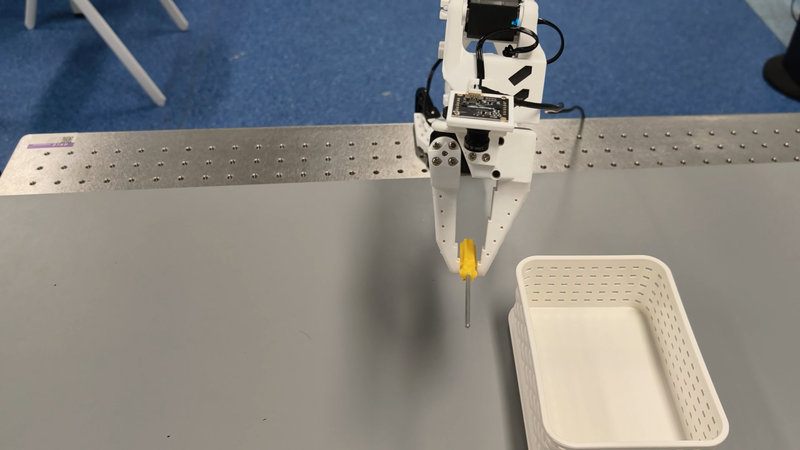}
        & \includegraphics[width=0.17\textwidth]{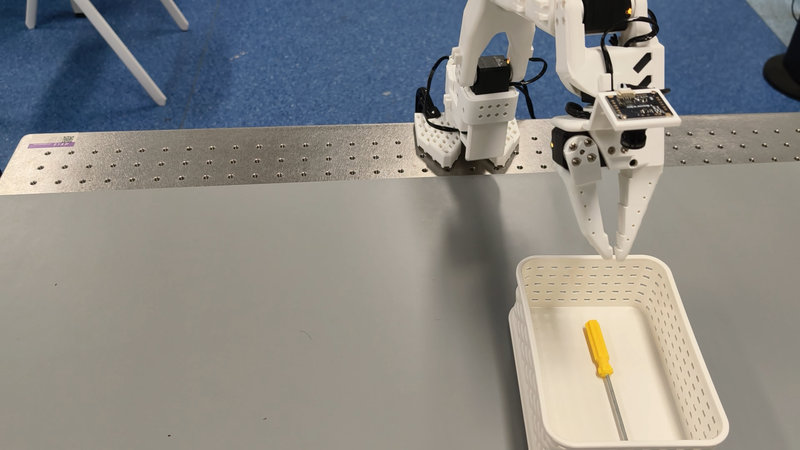} \\
        \taskrowlabel{Toy\\Duck}
        & \includegraphics[width=0.17\textwidth]{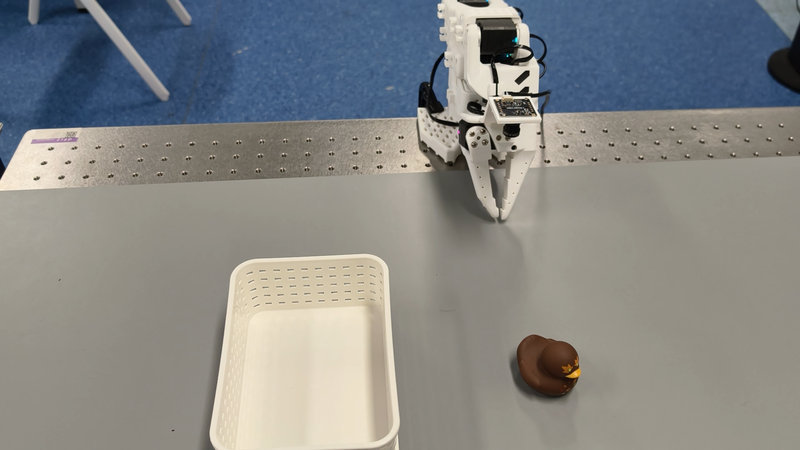}
        & \includegraphics[width=0.17\textwidth]{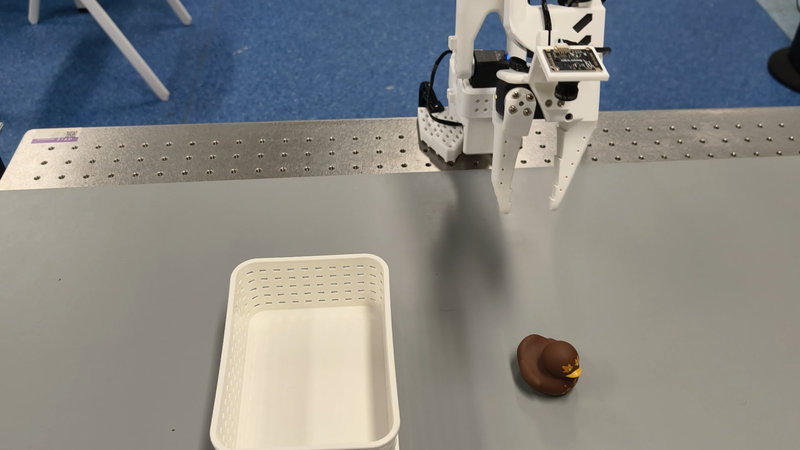}
        & \includegraphics[width=0.17\textwidth]{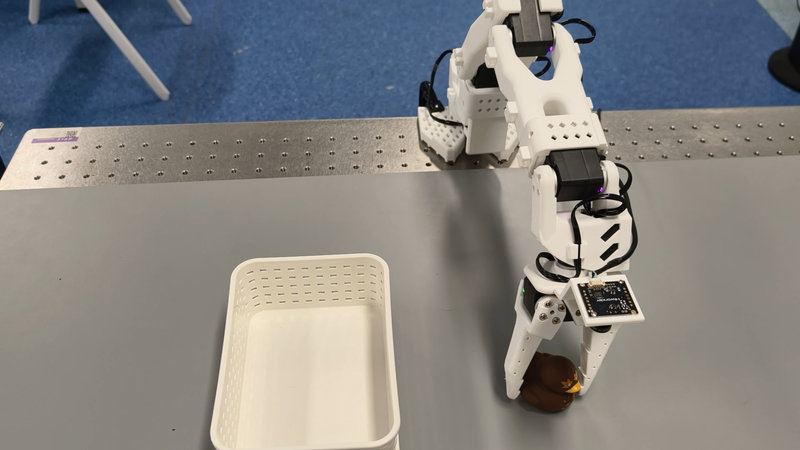}
        & \includegraphics[width=0.17\textwidth]{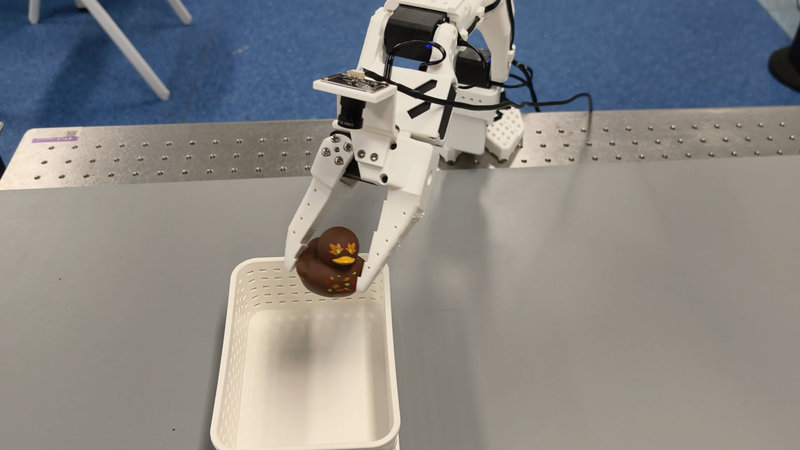}
        & \includegraphics[width=0.17\textwidth]{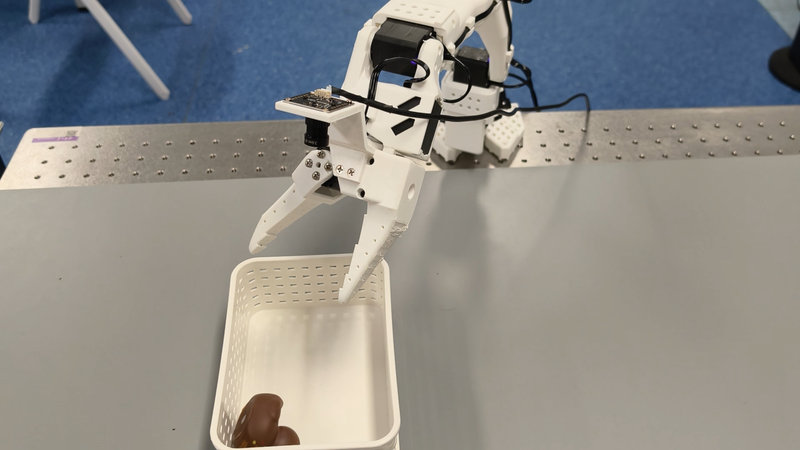} \\
        \taskrowlabel{Stack Bowls\\Two}
        & \includegraphics[width=0.17\textwidth]{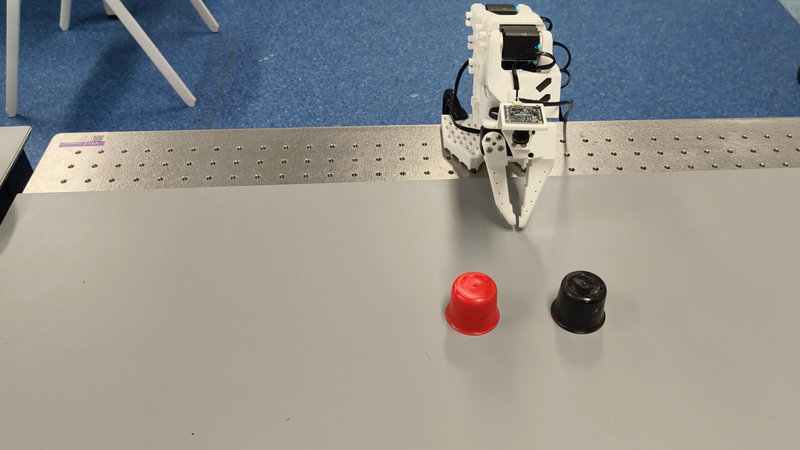}
        & \includegraphics[width=0.17\textwidth]{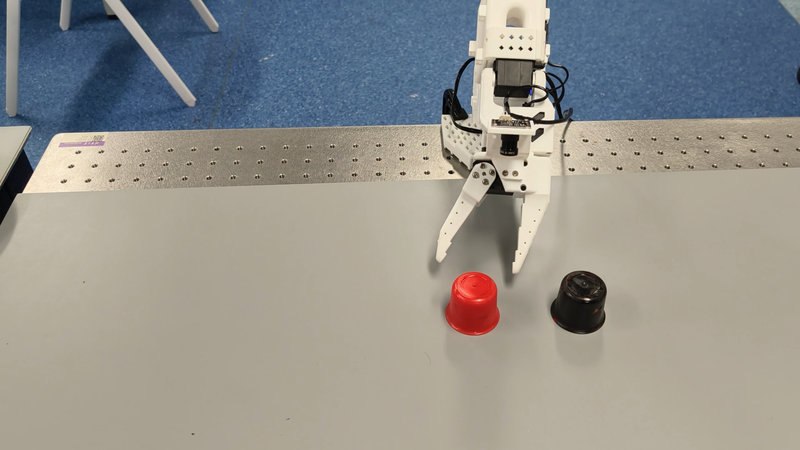}
        & \includegraphics[width=0.17\textwidth]{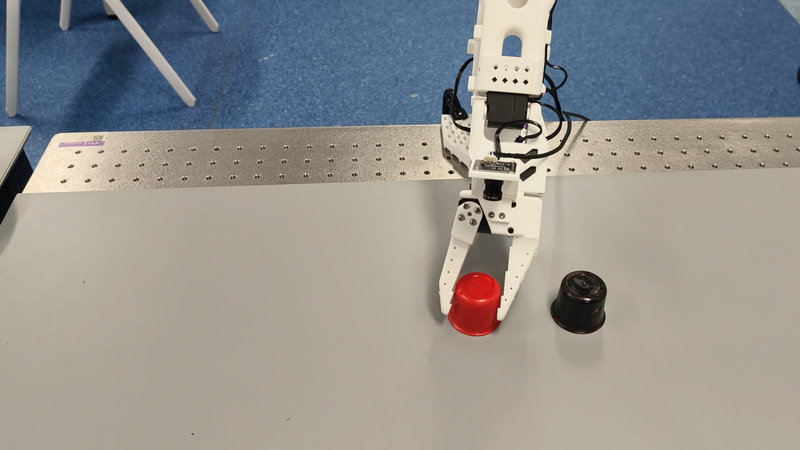}
        & \includegraphics[width=0.17\textwidth]{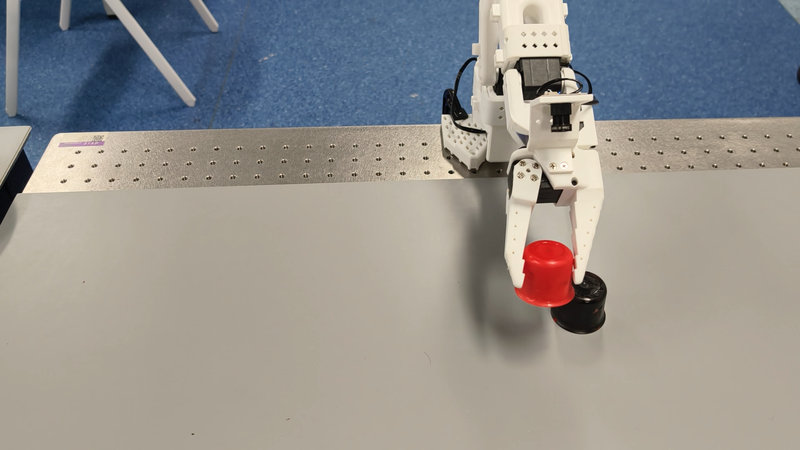}
        & \includegraphics[width=0.17\textwidth]{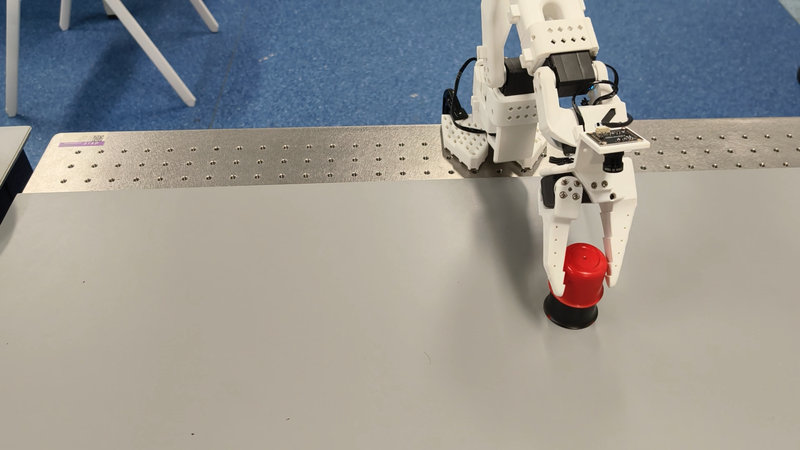} \\
        \taskrowlabel{Pour\\Water}
        & \includegraphics[width=0.17\textwidth]{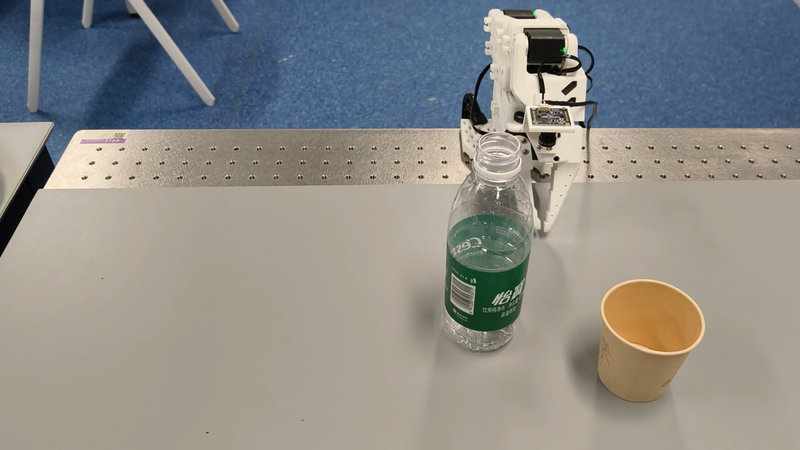}
        & \includegraphics[width=0.17\textwidth]{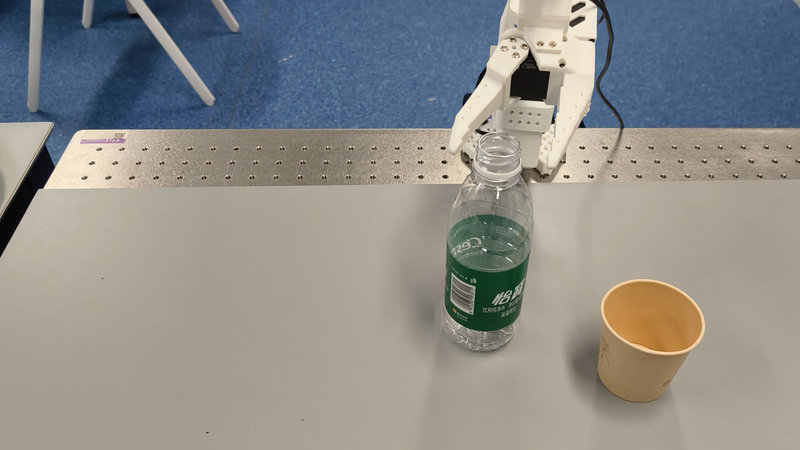}
        & \includegraphics[width=0.17\textwidth]{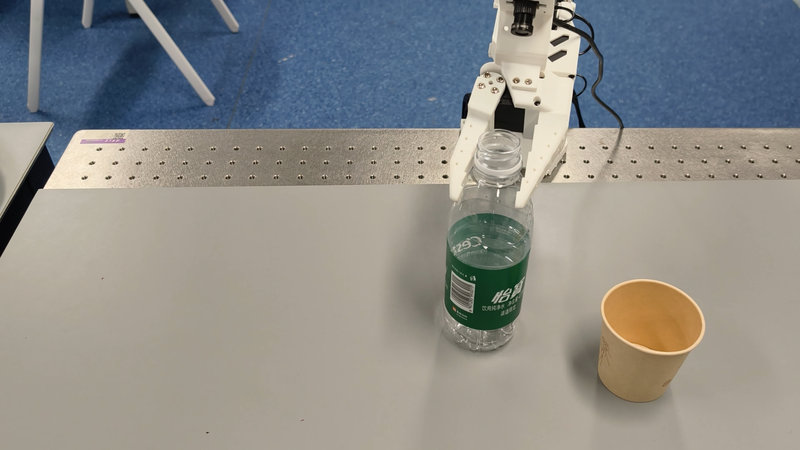}
        & \includegraphics[width=0.17\textwidth]{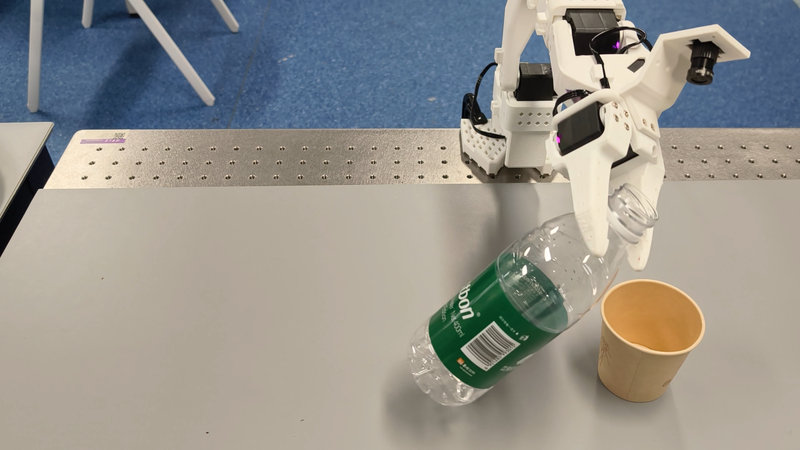}
        & \includegraphics[width=0.17\textwidth]{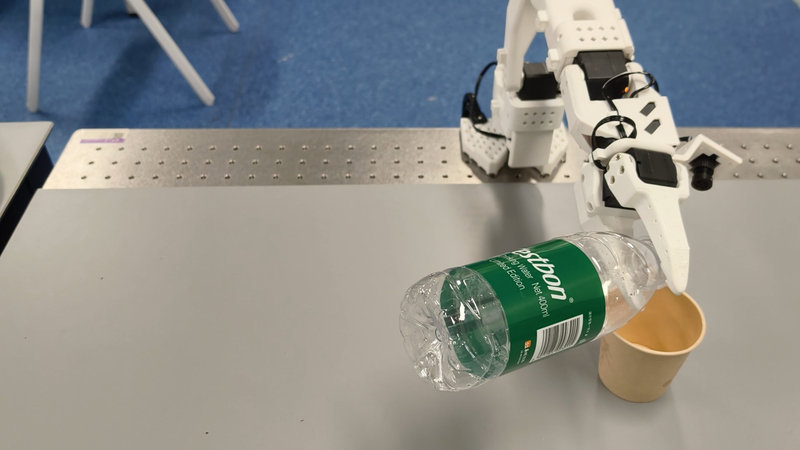}
    \end{tabular}
    \caption{Real-world manipulation rollouts. Columns show the temporal progression from $t_0$ to $t_4$, while rows correspond to the six tasks. To provide clearer visualizations than the low-resolution front camera used by the policy, these frames are captured by an auxiliary camera placed alongside the front camera. Consequently, the robot arms are not centered in the images.}
    \label{fig:real_world_visualizations}
\end{figure*}

Figure~\ref{fig:real_world_visualizations} presents representative SURE rollouts for all six real-world manipulation tasks, illustrating the temporal progression of each task from the initial scene to the final state.

\end{document}